\documentclass{article}

    \PassOptionsToPackage{numbers, compress}{natbib}

\usepackage[preprint]{neurips_2026}

\usepackage[utf8]{inputenc} 
\usepackage[T1]{fontenc}    
\usepackage{hyperref}       
\usepackage{url}            
\usepackage{booktabs}       
\usepackage{amsfonts}       
\usepackage{nicefrac}       
\usepackage{microtype}      
\usepackage{xcolor}         

\usepackage{graphicx}
\usepackage{multirow}

\usepackage{wrapfig}
\usepackage[table]{xcolor}
\usepackage{makecell}   
\usepackage{soul}       
\usepackage{enumitem}   
\usepackage{etoc}
\title{BARISTA: A Multi-Task Egocentric Benchmark for Compositional Visual Understanding} 

\author{
{\bfseries
Patrick Knab\textsuperscript{1,2,*}\;
Orgest Xhelili\textsuperscript{1,*}\;
Inis Buzi\textsuperscript{1}\;
Drago Andres Guggiana Nilo\textsuperscript{1}
}\\[-0.2ex]
{\bfseries
Mohd Saquib Khan\textsuperscript{1}\;
Lorenz Kolb\textsuperscript{1}\;
Manuel Scherzer\textsuperscript{1}\;
Kerem Yildirir\textsuperscript{1}
}\\[-0.2ex]
{\bfseries
Christian Bartelt\textsuperscript{2}\;
Philipp Johannes Schubert\textsuperscript{1}
}\\[-0.2ex]
{\normalfont\footnotesize
\textsuperscript{1}\,Ramblr.ai Research
\quad
\textsuperscript{2}\,Technical University of Clausthal
\quad
\textsuperscript{*}\,Equal contribution.
}
}

\begin{document}
\etocdepthtag.toc{mtmain}

\maketitle

\begin{abstract}
Scene understanding is central to general physical intelligence, and video is a primary modality for capturing both state and temporal dynamics of a scene.
Yet understanding physical processes remains difficult, as models must combine object localization, hand-object interactions, relational parsing, temporal reasoning, and step-level procedural inference.
Existing benchmarks usually evaluate these capabilities separately, limiting diagnosis of \emph{why} models fail on procedural tasks.
We introduce \textbf{BARISTA}, a densely annotated egocentric dataset and benchmark of $185$ real-world coffee-preparation videos covering fully automatic, portafilter-based, and capsule-based workflows.
BARISTA provides verified per-frame scene graphs linking persistent object identities to masks, tracks, boxes, attributes, typed relations, hand-object interactions, activities, and process steps.
From these graphs, we derive zero-shot language-based tasks spanning phrase grounding, hand-object interaction recognition, referring, activity recognition, relation extraction, and temporal visual question answering.
Experiments reveal strong variation across task families and no consistently dominant model family, positioning BARISTA as a challenging diagnostic benchmark for procedural video understanding. Code and dataset available at \href{https://huggingface.co/datasets/ramblr/BARISTA}{https://huggingface.co/datasets/ramblr/BARISTA}.
\end{abstract}

\begin{figure}[h]
    \centering
    \includegraphics[width=\linewidth]{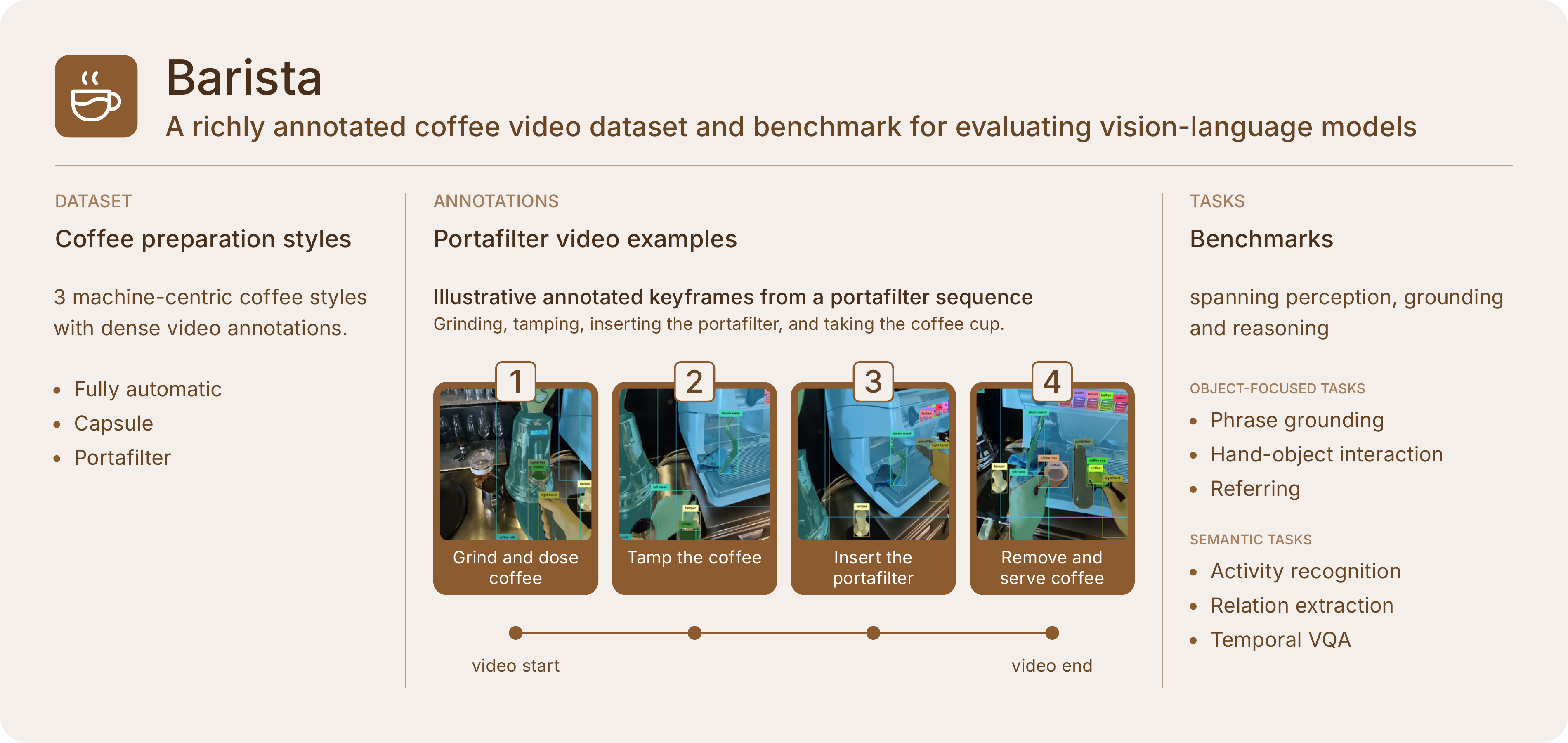}
    \caption{\textbf{BARISTA overview.} BARISTA is an egocentric coffee-preparation dataset and benchmark spanning three preparation styles, dense task-relevant annotations, and object-focused and semantic evaluation tasks derived from a shared scene graph.}
    \label{fig:intro-figure}
\end{figure}

\section{Introduction}\label{sec:intro}
\textbf{VLM evaluation remains fragmented.} Vision-language models (VLMs) increasingly serve as the perception backbones of agentic and embodied systems operating in the physical world~\citep{driess2023palme,kim2024openvla, gao2024physically, assran2025v, team2023gemini, ahn2022can}.
Such systems require a coordinated set of capabilities: localization of task-relevant objects and regions, interpreting interactions with the environment, recognizing ongoing activities in context, and reasoning about how the scene evolves over time \citep{team2023gemini,intelligence2025pi,knab2025concepts, cho2025perceptionlm}.
These abilities span multiple levels of abstraction, from masks and boxes to typed relations and process-level reasoning, and are difficult to assess when each capability is benchmarked on a different dataset under its own assumptions and evaluation protocol.
This points to a core benchmark-design requirement: evaluations must support controlled, fine-grained analysis of failure across levels, not merely report aggregate scores.
Most benchmarks isolate a single aspect of compositional understanding, such as activity recognition, long-video question answering, or object-centric perception~\citep{mangalam2023egoschema,li2023mvbench,fu2024videomme,salehi2024actionatlas}.
As a result, failures are hard to interpret: errors on process-level tasks may stem from weak localization, interaction understanding, relational reasoning, or temporal modeling, without a way to disentangle them.
This is particularly problematic given recent evidence that VLMs remain brittle even on basic spatial cognition~\citep{ahn2022can, khemlani2025vision,fu2024blink, prasse2026codebooks}.
Existing egocentric and instructional datasets offer a path to tackling this---providing dense spatial annotations, activity labels, or relational structure---but each typically covers only part of the required supervision. 
Because their domains, label spaces, and evaluation setups differ, they provide limited common ground to trace a high-level reasoning failure back to a concrete perceptual gap.

\textbf{This work.} We introduce \textsc{Barista}, a densely annotated egocentric dataset and unified benchmark of $185$ real-world coffee-preparation videos.
Its core is a per-frame co-registered scene graph that links masks, boxes, attributes, typed relations, hand-object interactions, and procedural labels through persistent object identities.
Across $4.4$ hours, BARISTA contains $3.61\mathrm{M}$ instance masks, $4.33\mathrm{M}$ attributes, and $2.48\mathrm{M}$ typed relations.
It is intentionally compact but comprehensive: more of a controlled diagnostic testing environment than a large-scale resource.
This shared representation makes object-, interaction-, relation-, and process-level tasks directly comparable, enabling matched analysis of failures across levels.

\textbf{Benchmark and findings.} Figure \ref{fig:intro-figure} overviews the preparation styles, annotation layers, and tasks.
BARISTA supports phrase grounding, hand-object interaction recognition, segmentation, referring expression generation, activity recognition, relation extraction, and temporal visual question answering.
We evaluate language-facing tasks under a shared zero-shot protocol; their graph-derived construction makes process-level errors traceable to underlying localization, interaction, and relation failures.
Across Qwen \citep{yang2025qwen3}, Gemma \citep{team2024gemma}, GPT \citep{achiam2023gpt}, Claude \citep{anthropic2026claudeopus55}, and Gemini \citep{team2023gemini}, no model performs uniformly best: spatial grounding varies substantially more than semantic recognition, and localization-heavy capabilities are largely orthogonal to those driving temporal QA.

Our main contributions are as follows:
\begin{itemize}[nosep]
    \item We introduce \textsc{Barista}, a densely annotated egocentric dataset of $185$ coffee-preparation videos built around a per-frame co-registered scene graph linking masks, boxes, attributes, relations, hand-object interactions, and procedural steps via persistent object identities.
    \item We propose a semi-automated annotation pipeline and a graph-based task-generation procedure for dense spatial, relational, interaction, and language-derived supervision.
    \item We define a unified, decomposable benchmark with standardized zero-shot prompting and task-specific metrics, and use it to show that current frontier and open-weight VLMs exhibit \emph{task-orthogonal} capability profiles.
\end{itemize}

\section{Related work}\label{sec:related_work}

\definecolor{coffeebrown}{RGB}{232, 220, 200}
\newcommand{\cmark}{\checkmark}

\textbf{Egocentric and instructional video datasets.}
Large-scale egocentric resources such as EPIC-KITCHENS~\citep{Damen_2018_ECCV}, Ego4D~\citep{grauman2022ego4d}, Ego-Exo4D~\citep{grauman2024ego}, and the long-form EgoLife~\citep{yang2025egolife} have substantially advanced first-person video understanding, primarily through narration-derived action labels and temporal boundaries rather than per-frame structured annotations.
VISOR~\citep{darkhalil2022epic} extends EPIC-KITCHENS with $272\mathrm{K}$ manual pixel masks for hands and active objects across $257$ classes, $9.9\mathrm{M}$ interpolated dense masks, and $67\mathrm{K}$ hand-object contact-state relations, but does not annotate typed semantic relations between objects, per-instance attribute vocabularies, or activity segments.
\begin{table}[t]
\centering
\caption{\textbf{Comparison of selected related video datasets.}
\textit{sparse}: objects annotated only at movement events.
\textit{contact}: hand-object contact only.
\textit{fixture}: object-fixture associations from 3D lifting.
\textit{typed}: semantic object-object or hand-object relation labels beyond boxes/masks.}
\resizebox{\textwidth}{!}{%
\setlength{\tabcolsep}{4pt}
\renewcommand{\arraystretch}{1.2}
\begin{tabular}{l c c c c c c c c}
\toprule
\textbf{Dataset} &
\textbf{View} &
\makecell{\textbf{Dataset}\\\textbf{Size}} &
\makecell{\textbf{Instance}\\\textbf{Masks}} &
\makecell{\textbf{Bounding}\\\textbf{Boxes}} &
\makecell{\textbf{Object}\\\textbf{Attributes}} &
\makecell{\textbf{Typed}\\\textbf{Relations}} &
\makecell{\textbf{Action}\\\textbf{Labels}} &
\makecell{\textbf{Temporal}\\\textbf{Tracks}} \\
\midrule
EPIC-KITCHENS \citep{Damen_2018_ECCV}     & Ego     & 100h     &            &            &            &            & \cmark     &            \\
VISOR (EPIC) \citep{darkhalil2022epic}     & Ego     & 36h      & \cmark     & \cmark     &            & contact    &            & \cmark     \\
HD-EPIC \citep{Perrett_2025_CVPR}          & Ego     & 41h      & sparse     & \cmark     &            & fixture    & \cmark     & \cmark     \\
Ego4D \citep{grauman2022ego4d}             & Ego     & 3000h    &            & \cmark     &            &            & \cmark     &            \\
Ego-Exo4D \citep{grauman2024ego}           & Ego+Exo & 1286h    &            & \cmark     &            &            & \cmark     &            \\
Assembly101 \citep{sener2022assembly101}   & Ego+Exo & 513h     &            &            &            &            & \cmark     &            \\
EPFL-SmartKitchen-30 \citep{bonetto2025epfl}     & Ego+Exo & 29.7h    &            &            &            &            & \cmark     &            \\
PVSG \citep{Yang_2023_CVPR}               & Ego+Exo & 9h & \cmark     & \cmark     &            & \cmark     &            & \cmark     \\
\midrule
\rowcolor{coffeebrown}
\textbf{BARISTA (Ours)} & Ego & 4.4h &
\textbf{\cmark} & \textbf{\cmark} & \textbf{\cmark} & \textbf{\cmark} & \textbf{\cmark} & \textbf{\cmark} \\
\bottomrule
\end{tabular}
}
\label{tab:dataset_comparison}
\end{table}
HD-EPIC~\citep{Perrett_2025_CVPR} is the closest recent point of comparison, providing $41.3$ hours of unscripted kitchen recordings with dense hand masks, sparse object masks, 3D digital twins, gaze, audio, and a $26\mathrm{K}$-question VQA benchmark.
However, object masks are temporally sparse, and the only explicitly annotated relation type connects objects to kitchen fixtures via 3D lifting; other inter-object and hand-object relations must be recovered from narration parsing.
Egocentric procedural datasets provide complementary supervision.
Assembly101~\citep{sener2022assembly101} and HOI4D~\citep{liu2022hoi4d} offer dense 3D hand pose and multi-granularity activity segments, with HOI4D adding panoptic segmentation and 6-DoF object pose.
HoloAssist~\citep{wang2023holoassist} and EPFL-Smart-Kitchen-30~\citep{bonetto2025epfl} capture rich multi-modal signals (gaze, dialogue, 3D body, and hand pose) for interactive assistance and fine-grained cooking.
Instructional benchmarks such as YouCook2~\citep{zhou2018towards}, COIN~\citep{tang2019coin}, and EgoProceL~\citep{bansal2022egoprocel} provide temporal step boundaries at scale but lack per-frame spatial annotations.
Scene-graph datasets such as Action Genome~\citep{ji2020actiongenome}, PVSG~\citep{Yang_2023_CVPR}, and AGQA~\citep{grundemclaughlin2021agqa} offer relational supervision but outside of procedural egocentric settings.

\textbf{VLM benchmarking.}
Recent video-VLM benchmarks evaluate complementary aspects of video understanding: EgoSchema~\citep{mangalam2023egoschema} and LongVideoBench~\citep{wu2024longvideobench} test long-form temporal reasoning over minutes to hours, MVBench~\citep{li2023mvbench} and Video-MME~\citep{fu2024videomme} provide broad multi-task or multi-domain coverage, and ActionAtlas~\citep{salehi2024actionatlas} and FAVOR-Bench~\citep{tu2025favor} target fine-grained action or motion discrimination.
Together, they span a diverse evaluation landscape from temporal localization and action discrimination to broad multi-domain video comprehension. 
However, these benchmarks typically evaluate final-answer accuracy rather than grounding each question in a verified per-frame scene graph.
Closest in diagnostic intent are VSTAT~\citep{yu2026vstat}, which separates perceptual from reasoning limitations via text transcriptions of the videos, and Flat-Pack Bench~\citep{chetan2026flatpack}, which categorizes errors from models' reasoning summaries. 
BARISTA instead scores perceptual and process-level tasks on the same events against verified annotations.

\textbf{Positioning.}
Table \ref{tab:dataset_comparison} summarizes the datasets most directly comparable to BARISTA.
BARISTA is smaller in scale ($185$ videos, $\sim\!469\mathrm{K}$ annotated frames) but provides dense per-frame co-registered supervision: $3.61\mathrm{M}$ instance masks across $46$ object categories, $2{,}424$ activity segments over $108$ classes.
Each of these layers has precedents, but prior resources do not co-register them on the same frames through persistent object identities.
Every benchmark task is derived deterministically from this verified scene graph, and each item is grounded in verified instance masks, relations, and activity segments, so that model failures can be diagnosed at the spatial and interaction level.
Unlike broad video leaderboards that primarily aggregate end-task scores, BARISTA is designed as a controlled, decomposable benchmark whose task families help expose failures across perceptual and reasoning levels, including matched analyses of the same events (Section~\ref{sec:failure_analysis}).
\newpage
\section{BARISTA}\label{sec:dataset}

\subsection{Dataset creation}\label{sec:dataset_collection}
\textbf{Coffee preparation tasks.}
Coffee preparation is a compact but expressive procedural domain.
Different preparation styles share the same high-level goal while varying substantially in required tools, visible machine state, hand usage, step order, and interaction duration.
BARISTA focuses on three representative preparation types:
\begin{itemize}[nosep]
    \item \textit{Fully automatic machines}, which emphasize short but visually diverse interactions such as button pressing, milk handling, cup placement, and machine-state feedback.
    \item \textit{Capsule machines}, which involve inserting a capsule, opening and closing the lever, and automated extraction with substantial variation in the machine form factor.
    \item \textit{Portafilter-based espresso preparation}, which captures a substantially more manual workflow, including dosing, tamping, attaching the portafilter, and starting extraction.
\end{itemize}
Together, these procedures span both fast interface-driven actions and longer manipulation-heavy sequences, making them useful for studying perception, grounding and reasoning in one domain.

\textbf{Data collection procedure.}
All $185$ videos were recorded in-house by people affiliated with the research group in controlled real-world coffee-preparation setups. 
No external participants or crowdworkers were involved in data capture.
Recording devices include smartphones and smart glasses (Apple Vision Pro, RayBan Meta~3, RayBan Wayfarer), so that the same procedural domain is captured across heterogeneous egocentric hardware.
The recorders varied tools, machines, coffee variants, and optional steps such as grinding or milk preparation, which introduces natural procedural variation without changing the underlying domain.
In total, $10$ recorders captured the videos in $14$ distinct locations on $15$ distinct coffee machines (Table~\ref{tab:diversity}, Appendix~\ref{app:diversity}).

\subsection{Annotation pipeline}\label{sec:dataset_pipeline}
\begin{figure}[t]
    \centering
    \includegraphics[width=\linewidth]{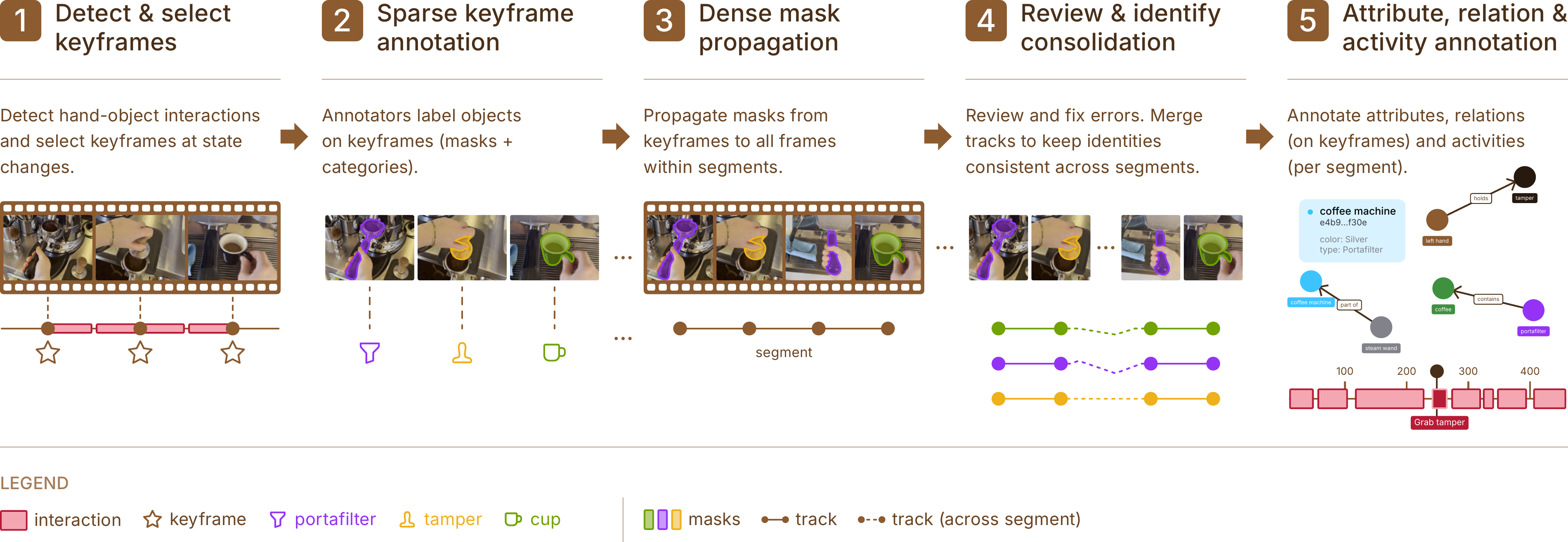}
    \caption{\textbf{BARISTA annotation pipeline.} Egocentric coffee-preparation videos are reduced to interaction segments and keyframes, sparsely annotated with object masks and categories, densely propagated across frames, reviewed for identity consistency, and enriched with attributes, typed relations, activities, and procedural step labels. This produces a per-frame co-registered scene graph linking spatial, relational, interaction, and temporal supervision.}
\label{fig:annotation-pipeline}
\end{figure}

Producing temporally consistent supervision for egocentric manipulation videos is challenging because hands frequently occlude objects, camera motion is substantial, and meaningful actions can be brief.
To scale BARISTA while preserving quality, we build a per-frame co-registered scene graph with a semi-automated pipeline that alternates model-assisted bootstrapping with targeted human verification~\citep{chen2026action100m, chen2024makes}.
Annotations were performed by trained annotators affiliated with the authors' institutions and human-reviewed: models only propose and propagate masks, and no annotation is generated by an LLM.
Figure \ref{fig:annotation-pipeline} visualizes the annotation pipeline described in the following. 

\textbf{Step~1: Interaction detection and keyframe selection.}
We first run a CaRe-Ego model~\citep{su2024careego}, finetuned on an internal egocentric hand-object interaction dataset used only for annotation bootstrapping, on every frame to produce per-frame interaction confidence scores. 
We extract interaction segments via temporal smoothing and morphological filtering of the raw signal (median filtering and binarization, connected-components extraction, removal of short segments, and merging of close gaps).
Within each interaction segment, we select its temporal center as a keyframe while maintaining a maximum inter-keyframe distance of 500 frames.

\textbf{Step~2: Sparse keyframe annotation.}
Annotators label all scene-relevant objects on the selected keyframes using a custom interface with interactive mask suggestions from SAM~2~\citep{ravi2024sam2}.
They can accept a proposed mask or refine it with brush and polygon tools.
Each instance receives a mask and category label and previously annotated objects are re-presented within the same segment to encourage identity reuse.

\textbf{Step~3: Dense mask propagation.}
We expand the sparse keyframe annotations into dense per-frame masks using the video propagation capabilities of SAM~2~\citep{ravi2024sam2} and SAM~3~\citep{sam3}.
For each propagation segment, the model is conditioned on all annotated keyframe masks simultaneously and propagates bidirectionally, so that intermediate frames benefit from reference appearances on both sides. 
This multi-frame conditioning provides greater robustness to occlusion, motion blur, and appearance change compared to single-frame propagation. 
Additionally, adjacent segments are propagated with shared temporal overlap, linking object identities across segment boundaries based on mask agreement.

\textbf{Step~4: Review and identity consolidation.}
Annotators review the propagated annotations in a correction interface to fix low-quality masks, missing masks after occlusion, or identity inconsistencies. 
They correct only sparse frames and repropagate within the problematic segments, keeping review efficient. 
Annotators then consolidate object identities across segments: when the same physical object reappears after occlusion or across neighboring segments, its per-frame sequences are merged into a single consistent identity.

\textbf{Step~5: Attribute, relation, and activity annotation.}
We complete the scene graph with three segment-level layers (object attributes, directed relations, and activities). 
Annotation is guided by a controlled domain vocabulary that is initialized before annotation and extended only through curator-approved additions when genuinely new entities appear.
Object-level attributes are assigned over each instance's active frame range. 
Directed typed relations are labeled on representative frames and propagated across a given segment, distinguishing spatial relations between objects (e.g. on, inside, part of) from hand-object actions (e.g. holds, presses, touches).
Lastly, activities are annotated at two granularities: fine-grained verb+noun labels (e.g. place cup) and coarser procedural steps reflecting the high-level process structure.

\subsection{Dataset analysis}\label{sec:dataset_analysis}
\textbf{Statistics.} BARISTA comprises $185$ videos and $469{,}201$ annotated frames with a dense graph-structured annotation layer: $3.61\mathrm{M}$ instance masks with bounding boxes across $46$ object categories, $4.33\mathrm{M}$ per-instance attribute annotations (color, type, state), $2.48\mathrm{M}$ typed relations ($12$ predicates covering spatial arrangement and hand-object interactions), $2{,}424$ fine-grained activity segments spanning $108$ classes, and $1{,}305$ process steps across $13$ classes.
Videos are exported at $30$ FPS across heterogeneous devices and resolutions ranging from $1280{\times}720$ to $1920{\times}1080$.
Annotated frames contain on average $7.63$ objects (median $7$, up to $24$) and $5.3$ typed relations per frame.
A detailed distributional analysis, including per-category frequencies, activity class counts, relation predicate distributions, and duration statistics, is provided in Appendix \ref{app:dataset_stats}.
We also include an 80/20 video-level train/test split, stratified by preparation type, with $148$/$37$ videos, that is intended for fine-tuning.

\textbf{Annotation Quality.}
Because re-annotating every video multiple times is impractical, we repeat the five-stage annotation pipeline on six randomly selected videos, which are independently re-annotated by two annotators.
The quality study covers 58 original keyframes and 240 additional manually re-annotated frames, no further propagated frames are evaluated. 
Of the additional frames, 75 were selected to improve object-category coverage, while 165 were sampled at temporal offsets from keyframes to directly evaluate propagated annotations.
Across all pairwise comparisons, mean mask IoU was $0.940$ for existing keyframes and $0.932$ for additional frames, category accuracy was $0.998$, and mean F1 was $0.998$ for attributes, $0.999$ for relations, and $0.989$ for activities, indicating high annotation quality, reliable propagation, and strong inter-annotator agreement.
\section{BARISTA benchmark}\label{sec:benchmark}
\subsection{Unified evaluation protocol}\label{sec:benchmark_protocol}
The BARISTA benchmark defines a suite of zero-shot tasks that probe complementary aspects of egocentric procedural understanding.
All tasks are derived deterministically from the verified scene graph described in Section \ref{sec:dataset}, ensuring that each evaluation example is traceable to the same underlying annotations.
Depending on the task, examples use either single-frame input (phrase grounding, hand-object interaction recognition, referring expression generation, and relation extraction) or multi-frame clips (activity recognition and temporal visual question answering).
For each task, annotated frames or clips are paired with task-specific ground truth, including bounding boxes, interaction labels, relation tuples, or action labels.
At evaluation time, each example is rendered into a multimodal prompt, and the model's response is scored against the ground truth with a task-appropriate metric.

\textbf{Sampling.} 
To keep the evaluation cost manageable while preserving task diversity, we use fixed evaluation subsets in the main benchmark: $1000$ samples for single-frame tasks, $500$ samples for activity recognition with $4$ frames per example, and $300$ samples for temporal VQA with $6$ frames per example.
The subsets are sampled from all $185$ videos (zero-shot; Section~\ref{sec:dataset_analysis}).
Because samples are deterministically derived from dense annotations, users can scale benchmarks, adjust sampling criteria, or run standardized ablations on shared subsets.
This scalability also supports fine-grained error analysis: users can sample additional instances from critical video segments and inspect model behavior in regions of interest.

\textbf{Models.}
Our main evaluation covers two closed models, Gemini~3~Flash~\citep{team2023gemini} and GPT-5.4~mini~\citep{achiam2023gpt}, and three open-weight models, Qwen3.5-27B, Qwen3.5-9B~\citep{yang2025qwen3}, and Gemma~4-31B~\citep{team2024gemma}.
Each task specifies a task-specific system prompt and a user-prompt template, and these prompts are shared across models with minimal provider-specific adaptations (e.g.\ bounding-box format).
For closed-source models, we set the thinking effort to low and query with temperature $0.7$, except for GPT-5.4~mini, which does not allow for custom sampling.
To cover stronger frontier models and a higher thinking budget, we additionally evaluate Gemini~4~Argon, Gemini~4~Hydrogen, and Claude~Opus~5.5 with medium thinking effort on the same frozen subsets (Appendix~\ref{app:benchmark_conventions}).
We run the evaluations three times and report the mean and standard deviation.
Two specialist baselines complement the VLMs: CaRe-Ego~\citep{su2024careego} for egocentric hand-object interaction, and SAM~3~\citep{sam3} for open-vocabulary grounding.
Each example includes its annotations for reproducible, provider-agnostic evaluation across hosted APIs and open-weight models served via vLLM~\citep{vllm}.

\begin{wraptable}{r}{0.42\textwidth}
\centering
\vspace{-1.em}
\caption{\textbf{Phrase grounding.} COCO-style mAP on $1000$ frozen single-frames.}
\vspace{0.5em}
\small
\setlength{\tabcolsep}{2.5pt}
\begin{tabular}{@{}l cc@{}}
\toprule
Model & mAP@50 & mAP \\
\midrule
SAM 3          & $.393$ & $.362$ \\
\midrule
Gemini~3~Flash
    & $.697 \pm .017$
    & $.534 \pm .008$ \\
GPT-5.4~mini
    & $.163 \pm .022$
    & $.068 \pm .014$ \\
\midrule
Gemini~4~Argon
    & $\mathbf{.770} \pm .004$
    & $\mathbf{.629} \pm .002$ \\
Gemini~4~Hydrogen
    & $.742 \pm .006$
    & $.603 \pm .003$ \\
Claude~Opus~5.5
    & $.751 \pm .006$
    & $.579 \pm .009$ \\
\midrule
Gemma 4-31B
    & $.634 \pm .005$
    & $.432 \pm .004$ \\
Qwen3.5-27B
    & $.680 \pm .016$
    & $.518 \pm .005$ \\
\midrule
Qwen3.5-9B
    & $.559 \pm .027$
    & $.350 \pm .020$ \\
\bottomrule
\end{tabular}
\vspace{-3em}
\label{tab:grounding_results}
\end{wraptable}

\subsection{Task families with benchmarks}
The benchmark is organized into two task families that mirror the decomposition introduced in Section \ref{sec:intro}. 
Object-based perception tasks---phrase grounding, egocentric HOI recognition, and referring expression generation---are shown in Section~\ref{object_tasks}, while semantic scene understanding tasks---activity recognition, relation extraction, and temporal VQA---are shown in Section~\ref{semantic_tasks}.
This separation reflects BARISTA's central design goal: tasks are derived from the same verified per-frame scene graph, but probe different abstraction levels, from grounded object and interaction understanding to relational, procedural, and temporal reasoning.

\label{sec:task_definitions}

\subsubsection{Object-focused tasks}
\label{object_tasks}

\textbf{Phrase grounding}~\citep{liu2024groundingdino,you2023ferret} evaluates whether a model can link a short natural-language phrase, such as \textit{``the red button on the automatic coffee machine''} to the correct object in a coffee-making scene.
Given a single frame, the model must predict a bounding box for the referenced object.
Phrases are constructed from verified category, color, and spatial-relation annotations, selecting the richest available description for each category, which makes them context-dependent and rarely reducible to a simple category label.
Performance is measured using COCO-style~\citep{lin2014coco} mean average precision (mAP): mAP@50 measures detection quality at a loose IoU threshold of $0.5$ (correct region), while mAP averages over IoU thresholds from $0.5$ to $0.95$ and rewards tight box localization.
In Table \ref{tab:grounding_results}, we compare zero-shot VLMs with SAM~3 as an open-vocabulary grounding specialist on the same frozen evaluation set. 
Gemini~3~Flash and Qwen3.5-27B lead among the five main models; the three frontier models reach $.579$--$.629$ mAP.
SAM~3 is competitive but remains below the strongest VLMs, whereas GPT-5.4~mini performs poorly in this setting---it is included as a general-purpose VLM reference but is not primarily optimized for region-conditioned visual grounding.
A per-size-bin breakdown (Appendix~\ref{app:grounding}) shows that performance drops substantially on medium-sized objects, indicating that object scale remains a key bottleneck.

\textbf{Egocentric hand-object interaction recognition}~\citep{su2024careego,zhang2022egohos} requires the model to detect and characterize all hand-object interactions visible in a single egocentric frame.
The ground truth specifies the manipulated object, the involved hand(s), and a hand-type label (left, right, or both).
The model must localize both
\begin{wraptable}{r}{0.52\textwidth}
    \centering
    \caption{\textbf{Egocentric hand-object interaction.} Interaction F1 and localization IoU on $1000$ frozen frames (IoU$\geq$0.5 matching).}
    \vspace{0em}
    \small
    \resizebox{\linewidth}{!}{%
    \setlength{\tabcolsep}{4pt}
    \begin{tabular}{@{}l ccc@{}}
        \toprule
        Model & Int. F1 & Hand IoU & Obj. IoU \\
        \midrule
        CaRe-Ego & $.490$ & $.703$ & $.839$ \\
        \midrule
        Gemini~3~Flash & $.749 \pm .008$ & $.744 \pm .007$ & $.892 \pm .001$ \\
        GPT-5.4~mini   & $.193 \pm .014$ & $.483 \pm .007$ & $.644 \pm .008$ \\
        \midrule
        Gemini~4~Argon$^\dagger$ & $.817 \pm .003$ & $\mathbf{.802} \pm .004$ & $.894 \pm .002$ \\
        Gemini~4~Hydrogen$^\dagger$ & $\mathbf{.822} \pm .002$ & $.738 \pm .002$ & $\mathbf{.898} \pm .001$ \\
        Claude~Opus~5.5$^\dagger$ & $.806 \pm .002$ & $.747 \pm .002$ & $.884 \pm .001$ \\
        \midrule
        Gemma 4-31B  & $.758 \pm .003$ & $.707 \pm .001$ & $.857 \pm .001$ \\
        Qwen3.5-27B  & $.753 \pm .007$ & $.706 \pm .004$ & $.864 \pm .002$ \\
        \midrule
        Qwen3.5-9B   & $.642 \pm .008$ & $.640 \pm .003$ & $.787 \pm .005$ \\
        \bottomrule
    \end{tabular}%
    }
\label{tab:hoi-results}
\vspace{0em}
\end{wraptable}
the object and the hand regions via bounding boxes and classify the type of hand interaction (whether the left hand, right hand, or both hands are involved).
The predicted interactions are greedily matched to the ground-truth annotations by object-box IoU at a $0.5$ threshold.
On matched pairs, we report interaction-level F1 and mean hand and object IoU.
Precision, recall, no-detection rate, and hand-type classification accuracy are reported in Appendix~\ref{app:hoi}.
We evaluate CaRe-Ego as a baseline specialist model on the same frames.
Table \ref{tab:hoi-results} shows that the strongest VLMs outperform CaRe-Ego on interaction F1, while CaRe-Ego remains competitive in localization, especially for object boxes.
Among the five main VLMs, Gemma~4-31B, Qwen3.5-27B, and Gemini~3~Flash are statistically tied on interaction F1, while Gemini~3~Flash produces the most accurate hand and object boxes.
The frontier models raise interaction F1 to $.806$--$.822$, with Gemini~4~Argon giving the best hand boxes.

\begin{wraptable}{r}{0.33\textwidth}
    \centering
    \vspace{-1em}
    \caption{\textbf{Referring expression generation.} LLM-judged correctness (G-Eval, $[0,1]$) on $1000$ frozen examples.}
    \label{tab:referring_results}
    \vspace{0.5em}
    \setlength{\tabcolsep}{4pt}
    \small
    \begin{tabular}{@{}lc@{}}
        \toprule
        Model & Correct. \\
        \midrule
        Gemini~3~Flash  & $.760 \pm .003$ \\
        GPT-5.4~mini    & $.308 \pm .006$ \\
        \midrule
        Gemini~4~Argon$^\dagger$ & $\mathbf{.843} \pm .003$ \\
        Gemini~4~Hydrogen$^\dagger$ & $.834 \pm .002$ \\
        Claude~Opus~5.5$^\dagger$ & $.765 \pm .009$ \\
        \midrule
        Gemma 4-31B     & $.667 \pm .004$ \\
        Qwen3.5-27B     & $.622 \pm .010$ \\
        \midrule
        Qwen3.5-9B      & $.576 \pm .009$ \\
        \bottomrule
    \end{tabular}
    \vspace{-0em}
\end{wraptable}

\textbf{Referring / region description}~\citep{you2023ferret,guo2024regiongpt} is the inverse of phrase grounding: instead of mapping language to a region, the model must describe a given region in context.
Each example consists of a frame and a specific bounding box, and the model must generate a concise description that identifies the marked object through its category, visual attributes, and spatial or functional relations to surrounding objects.
Ground-truth data include the object's category, attributes, direct relations, and context about related objects.
A separate LLM judge (Gemini~2.5~Flash) scores each prediction against this structured reference using G-Eval~\citep{liu2023geval} under a single Correctness criterion on a $0$--$10$ rubric (rescaled to $[0,1]$).
Because LLM-based evaluation is not calibrated, absolute scores are best interpreted for relative model comparison rather than as ground-truth quality estimates (more in Appendix~\ref{app:referring}).
Table \ref{tab:referring_results} shows that Gemini~4~Argon performs best ($.843$), followed by Gemini~4~Hydrogen, Claude~Opus~5.5, and Gemini~3 Flash, with Gemma~4-31B the strongest open-weight model, followed by Qwen3.5-27B and Qwen3.5-9B.
GPT-5.4~mini lags substantially behind, mirroring its weak performance on phrase grounding due to its limitations in spatially grounded understanding. 
Model rankings of the five main models are stable when re-scored by a cross-family judge ($\tau = 1.0$; Appendix~\ref{app:judge_robustness}), confirming the ordering reflects task performance rather than judge-specific bias.

\begin{wraptable}{r}{0.33\textwidth}
    \centering
    \vspace{-1.2em}
    \caption{\textbf{Activity recognition.} $25$-way MCQ accuracy on $500$ clips ($4$ frames each).}
    \label{tab:activity_mcq_results}
    \setlength{\tabcolsep}{4pt}
    \small
    \resizebox{0.9\linewidth}{!}{%
    \begin{tabular}{lc}
        \toprule
        Model & Acc. \\
        \midrule
        Gemini~3~Flash & $.765 \pm .015$ \\
        GPT-5.4~mini   & $.631 \pm .007$ \\
        \midrule
        Gemini~4~Argon$^\dagger$ & $\mathbf{.952} \pm .005$ \\
        Gemini~4~Hydrogen$^\dagger$ & $.915 \pm .001$ \\
        Claude~Opus~5.5$^\dagger$ & $.860 \pm .002$ \\
        \midrule
        Gemma 4-31B    & $.684 \pm .003$ \\
        Qwen3.5-27B    & $.778 \pm .005$ \\
        \midrule
        Qwen3.5-9B     & $.716 \pm .014$ \\
        \bottomrule
    \end{tabular}
    }
    \vspace{-1em}
\end{wraptable}

\newpage
\subsubsection{Semantic tasks}\label{semantic_tasks}

\textbf{Activity recognition} Given four frames sampled uniformly from an annotated activity segment, the model must select the correct activity label from $K{=}25$ multiple-choice options drawn from the full activity vocabulary.
We choose $K{=}25$ to make the task non-trivial while keeping the option set interpretable (constructions detailed in Appendix~\ref{app:activity}).
The results in Table \ref{tab:activity_mcq_results} show that Qwen3.5-27B and Gemini~3~Flash are effectively tied at the top of the five main models; the frontier models reach $.860$--$.952$.
The Qwen models show particular strength on this task, with Qwen3.5-9B outperforming both Gemma~4-31B and GPT-5.4~mini despite its smaller scale.

\textbf{Relation extraction} evaluates whether a model can recover the relational structure of a scene, moving beyond object recognition to structured scene understanding \citep{krishna2017visual}.

\begin{wraptable}{r}{0.52\textwidth}
    \centering
    \vspace{-1.0em}
    \caption{\textbf{Relation extraction.} Micro F1 under exact $(s,t,r,v)$ multiset matching on $1000$ frozen frames.}
    \label{tab:relation_extraction_combined}
    \vspace{0.5em}
    \small
    \resizebox{\linewidth}{!}{%
    \setlength{\tabcolsep}{4pt}
    \begin{tabular}{@{}l ccc@{}}
        \toprule
        Model & Overall & \texttt{human\_actions} & \texttt{position} \\
        \midrule
        Gemini~3~Flash  & $.675 \pm .008$ & $.675 \pm .004$ & $.710 \pm .007$ \\
        GPT-5.4~mini    & $.565 \pm .008$ & $.534 \pm .011$ & $.569 \pm .008$ \\
        \midrule
        Gemini~4~Argon$^\dagger$ & $.811 \pm .002$ & $.698 \pm .005$ & $.819 \pm .002$ \\
        Gemini~4~Hydrogen$^\dagger$ & $.798 \pm .002$ & $\mathbf{.742} \pm .003$ & $.802 \pm .002$ \\
        Claude~Opus~5.5$^\dagger$ & $\mathbf{.823} \pm .000$ & $.730 \pm .006$ & $\mathbf{.829} \pm .001$ \\
        \midrule
        Gemma 4-31B     & $.751 \pm .002$ & $.673 \pm .012$ & $.758 \pm .001$ \\
        Qwen3.5-27B     & $.745 \pm .005$ & $.626 \pm .006$ & $.756 \pm .004$ \\
        \midrule
        Qwen3.5-9B      & $.613 \pm .004$ & $.573 \pm .010$ & $.630 \pm .005$ \\
        \bottomrule
    \end{tabular}%
    }
\vspace{-1em}
\end{wraptable}
The model receives a single frame with a set-of-mark~\citep{yang2023som} overlay labeling each annotated object with a numerical identifier, a legend mapping identifiers to categories, and the constrained vocabulary of valid relation types and values.
Given this input, the model must predict all directed relation tuples $(s, t, r, v)$---source object, target object, relation type, and value---that hold between the identified objects.
Predictions are evaluated against the verified ground-truth set via exact multiset matching: a predicted tuple counts as correct only if it matches a ground-truth tuple on all four fields. 
Because we provide object identities and categories, the task isolates relations  rather than object detection.
We report overall and per relation type F1 and full metrics in \ref{app:relation_extraction}.
In Table \ref{tab:relation_extraction_combined}, Gemma~4-31B achieves the best overall F1 among the five main models, narrowly ahead of Qwen3.5-27B; the frontier models reach $.798$--$.823$, with Claude~Opus~5.5 best.

\textbf{Temporal visual question answering} evaluates open-ended reasoning \citep{yu2019activitynet} over short temporal windows rather than single images.
We detect temporal changes between clip endpoints from structured annotations---attribute transitions, hand-contact sequences, spatial-relation shifts, and activity-label boundaries---and prompt Gemini~2.5~Flash~Lite to convert each detected change into a natural-language question-answer pair. 
The resulting pairs span
\begin{wraptable}{r}{0.57\textwidth}
    \centering
    \vspace{-0.5em}
    \caption{\textbf{Temporal VQA.} LLM-judged accuracy and completeness (G-Eval, $[0,1]$) on $300$ clips ($6$ frames, $4$\,s).}
    \label{tab:visual_qa_combined}
    \small
    \resizebox{\linewidth}{!}{%
    \setlength{\tabcolsep}{4pt}
    \begin{tabular}{@{}l ccc@{}}
        \toprule
        Model & Accuracy & Completeness & Mean \\
        \midrule
        Gemini~3~Flash   & $0.696 \pm .015$ & $0.640 \pm .008$ & $0.665 \pm .009$ \\
        GPT-5.4~mini     & $0.701 \pm .007$ & $0.675 \pm .008$ & $0.686 \pm .003$ \\
        \midrule
        Gemini~4~Argon$^\dagger$ & $0.791 \pm .006$ & $0.744 \pm .006$ & $0.766 \pm .006$ \\
        Gemini~4~Hydrogen$^\dagger$ & $\mathbf{0.799} \pm .004$ & $\mathbf{0.756} \pm .003$ & $\mathbf{0.777} \pm .001$ \\
        Claude~Opus~5.5$^\dagger$ & $0.713 \pm .020$ & $0.630 \pm .004$ & $0.670 \pm .008$ \\
        \midrule
        Gemma 4-31B      & $0.672 \pm .008$ & $0.629 \pm .009$ & $0.649 \pm .007$ \\
        Qwen3.5-27B      & $0.726 \pm .004$ & $0.680 \pm .005$ & $0.701 \pm .006$ \\
        \midrule
        Qwen3.5-9B       & $0.674 \pm .016$ & $0.641 \pm .017$ & $0.655 \pm .017$ \\
        \bottomrule
    \end{tabular}%
    }
\vspace{-0.5em}
\end{wraptable}
four categories: \emph{state change}, \emph{action transition}, \emph{relation change}, and \emph{activity progression}.
At inference time, the model receives six frames from a four-second window and one question and must produce a free-form answer.
A separate LLM judge (Gemini~2.5~Flash) scores each response with G-Eval~\citep{liu2023geval} on accuracy (factual correctness) and Completeness (coverage of reference changes).
As in referring, the absolute G-Eval scores reflect judge-specific calibration and are most informative for ranking models rather than measuring absolute answer quality.
Rankings are fully preserved under a 
cross-family GPT-5.4~mini judge on mean score and Completeness ($\tau = 1.0$), while a single adjacent accuracy swap occurs between Gemini~3~Flash and GPT-5.4~mini (Appendix~\ref{app:judge_robustness}).
Table \ref{tab:visual_qa_combined} shows that, among the five main models, Qwen3.5-27B has the highest mean and accuracy scores, while being tied with GPT-5.4~mini on completeness.
Both Gemini~4 models exceed this range ($.766$, $.777$); Claude~Opus~5.5 ($.670$) stays within it.
GPT-5.4~mini performs surprisingly well, ranking second on mean score despite weaker performance on other tasks.
Per-category results (Appendix~\ref{app:visual_qa}) show that relation-change questions are answered most reliably, whereas action transitions remain the most difficult category.

\subsection{Cross-task evaluation}
  \label{sec:failure_analysis}

  \begin{wraptable}{r}{0.48\textwidth}
  \centering
  \vspace{-1.2em}
  \caption{\textbf{Matched diagnostic cases.} Pass rates (\%) on $200$ cases shared across tasks.
``Trans.'' denotes activity-transition recognition.}
  \label{tab:failure_analysis}
  \scriptsize
  \setlength{\tabcolsep}{2.5pt}
  \begin{tabular}{@{}lrrrrrr@{}}
  \toprule
  Model & Ground. & HOI & Rel. & Act. & Trans. & Valid \\
  \midrule
  Qwen3.5-9B  & 47.0 & 72.3 & 67.8 & 80.4 & 94.4 & 86.5 \\
  Qwen3.5-27B & 53.9 & 77.1 & 83.1 & 82.8 & 95.5 & 95.2 \\
  Gemma 4-31B & 40.9 & 77.5 & 88.0 & 72.0 & 92.4 & 99.5 \\
  \bottomrule
  \end{tabular}
  \vspace{-1em}
  \end{wraptable}

Beyond evaluating tasks separately, we compare open-source VLMs on the same underlying video activity events. 
We construct 200 matched cases, each comprising an annotated event and its transition: grounding, hand-object interaction, and relation extraction use one anchor frame; activity recognition uses the surrounding segment; and activity-transition recognition uses an ordered clip from the same event. 
Matching outcomes across tasks lets us identify whether failures co-occur for the same event rather than on unrelated examples. 
For every model and case, we record whether each stage passes, fails, or produces an invalid output.
A stage passes when it meets its task-specific criterion: IoU for grounding, recall for hand-object
interaction, F1 for relations, and answer accuracy for activity and transition recognition.
Invalid outputs are not included in a stage's pass rate, but are included in the valid-output column.
The matched-case construction, task-specific pass criteria, conditional co-failure analysis, and case-level records are described in Appendix~\ref{app:failure_analysis}.
Table~\ref{tab:failure_analysis} shows clear differences between models.
Qwen3.5-27B has the highest pass rates for grounding, activity recognition, and transition recognition, while also producing valid outputs on most cases.
Gemma~4-31B performs best on relation extraction and is competitive on hand-object interaction, but is weaker on grounding and activity recognition.
Qwen3.5-9B has lower pass rates at every stage and produces more invalid outputs.

The co-failure analysis (Appendix~\ref{app:matched_cases}) shows that the link between perceptual and process-level errors is model-dependent: activity failures are significantly more likely after grounding or relation failures only for Qwen3.5-9B.
In an evidence-handoff intervention (Appendix~\ref{app:handoff}, Table~\ref{tab:handoff}), supplying ground-truth boxes, interactions, and relations of the anchor frame improves activity recognition by $+6.5$\,pp for Gemma~4-31B, the model with the weakest grounding, with no significant change for either Qwen model or for transition recognition. The benefit is thus model- and task-dependent.

\section{Discussion and conclusion}
\label{sec:discussion}

\paragraph{Key findings.}
BARISTA shows that holistic video understanding does not reduce to a single notion of ``multimodal capability'': models that excel on one part of the pipeline may still perform poorly on another and no model family dominates every task.
Gemini~3~Flash is strongest on object-focused tasks that depend on precise visual grounding---phrase grounding, referring expression generation, and accurate hand/object boxes for HOI---whereas the strongest open-weight models lead on structured prediction, with Gemma~4-31B achieving the best relation-extraction F1 and Qwen3.5-27B the best temporal VQA and activity-recognition performance.
This pattern shows that BARISTA separates localization, interaction recognition, relational parsing, activity recognition, and temporal reasoning.

A second consistent finding is that spatial grounding remains the main bottleneck.
Performance varies most on tasks requiring reliable region-level alignment, especially phrase grounding and HOI.
Even strong models degrade markedly on medium-sized objects: mAP drops by more than half between the large and medium size bins for every model, and the performance spread across models on grounding (mAP range $0.07$--$0.53$) is far wider than on semantic tasks such as activity recognition (accuracy range $0.63$--$0.78$) or temporal VQA (mean range $0.65$--$0.70$).
Current VLMs often capture the stage of a process before they can consistently localize the objects and interactions involved, which limits their ability to provide actionable feedback in physical workflows.

A third finding is that BARISTA exposes a meaningful internal structure within ``procedural understanding''.
Within temporal VQA, relation changes are answered most reliably, whereas action transitions remain hardest, suggesting that static before/after differences are easier than fine-grained transition dynamics.
This heterogeneity extends to the benchmark level: a post-hoc cross-task rank analysis (Appendix~\ref{app:decomposability}) shows that grounding, HOI, relations, and referring form a coherent cluster whose model rankings agree closely (Kendall $\tau \geq 0.6$ on $5$ of $6$ pairs), while temporal VQA is decoupled from this cluster ($\tau \in [-0.4, 0.0]$) and correlates only weakly with activity recognition ($\tau{=}0.4$).
Per-machine-type stratified results (Appendix~\ref{app:machine_type}) further show that the three preparation styles expose distinct performance profiles across tasks, confirming that procedural diversity is a meaningful evaluation axis.

\textbf{Limitations.}
BARISTA intentionally focuses on egocentric indoor coffee preparation, providing a controlled setting for compositional physical-world understanding rather than covering arbitrary domains.
Our annotations are dense over task-relevant entities and interactions rather than every visible object, reflecting this focus rather than open-world scene coverage.
Most remaining limitations concern the benchmark configuration in this paper rather than the BARISTA dataset itself.
Activity recognition uses multiple-choice classification with the correct answer among the candidates, enabling controlled comparison but likely yielding higher performance than open-vocabulary evaluation.
Parts of the benchmark use LLM-as-a-judge scoring, although a cross-family judge swap (Appendix~\ref{app:judge_robustness}) preserves model rankings ($\tau \geq 0.8$ on all metrics, $\tau = 1.0$ on three of four).
Finally, several evaluations use frozen subsets to control benchmarking cost, while the released annotations and sampling scripts allow straightforward expansion.

\textbf{Dataset utility beyond benchmarking.}
BARISTA is also designed to be useful as a training and analysis infrastructure.
Temporally consistent masks and tracks support segmentation and tracking; phrase-grounding pairs and boxes support localization; and relation tuples, HOI labels, and activity segments provide structured supervision for fine-tuning specialized perception modules.
Because all labels are linked through persistent object identities, the dataset can further support work on temporal scene graphs, interaction forecasting, process monitoring, failure analysis, and compositional debugging of vision-language systems.

\textbf{Conclusion.}
BARISTA advances compositional visual understanding by combining dense, task-relevant annotation with unified multi-task evaluation in a realistic egocentric procedural setting.
Its central contribution is diagnostic coherence and decomposability: the benchmark can trace a process-level failure back to localization, interaction recognition, relational structure, or temporal inference on the same underlying evidence.
Our experiments show that current VLMs remain uneven across these capabilities, with comparatively strong semantic recognition still falling short of consistently grounded procedural understanding.
By releasing videos, annotations, prompts, splits, and evaluation code, we provide a compact but extensible testbed for measuring not only whether procedural vision-language systems succeed, but where and possibly why they fail.

\section{Acknowledgments}
This work was partially funded by the Bavarian State Ministry for Economic Affairs, Regional Development and Energy (StMWi) under the Bavarian Collaborative Research Programs (BayVFP), funding line Digitalization / Information and Communication Technology, by the German Federal Ministry for Economic Affairs and Climate Action of Germany (BMWK), and in part by the German Federal Ministry for Research, Technology, and Space (BMFTR).

We thank Anna Lanz and Tooba Islam for leading the annotation process, and Christopher Zak for creating the paper figures and visual assets.

\bibliographystyle{unsrtnat}
\bibliography{main}


\newpage
\appendix
\etocdepthtag.toc{mtappendix}
\newpage

\etocsettagdepth{mtmain}{none}
\etocsettagdepth{mtappendix}{subsubsection}
\etocsettocstyle{\section*{Appendix contents}}{}
{\hypersetup{hidelinks}\tableofcontents}

\vspace{1em}

\section{Benchmark implementation details}
\label{app:benchmark_impl}

\subsection{Common conventions}
\label{app:benchmark_conventions}

\paragraph{Evaluation pipeline.}
Every task is evaluated on a deterministic, frozen set of examples derived from the released annotations.
For temporal VQA, reference question-answer pairs are generated once offline using Gemini~2.5~Flash Lite and cached so that downstream evaluation remains reproducible.
Each model is queried on the same frozen examples and scored with task-specific metrics.
For open-ended tasks, LLM-based judging is applied as a separate post-hoc step, allowing predictions to be re-scored under different judge configurations without repeating inference.

\paragraph{Bounding-box representation.}
Bounding boxes are stored canonically as absolute $(x_1,y_1,x_2,y_2)$ pixel coordinates in the frozen evaluation set.
At prompt construction time, coordinates are rescaled to the integer range $[0, 1000]$ and reordered into the format expected by each provider (e.g.\ $y_1, x_1, y_2, x_2$ for Gemini, Gemma and GPT, $x_1, y_1, x_2, y_2$ for Qwen).
This lets a single frozen dataset be reused across providers; model responses are always rescaled back to pixel coordinates before scoring.
The only exception is Claude~Opus~5.5 on phrase grounding and hand-object interaction: Anthropic's official documentation for Claude models advises against normalizing coordinates to $[0,1000]$, so to give the model a fair evaluation we prompt it for absolute pixel coordinates of the input frame and score its boxes directly, without rescaling; the ground truth, matching, and metrics are unchanged.

\paragraph{Decoding.}
All five main models are queried at temperature $0.7$, except GPT-5.4~mini, for which OpenAI does not expose custom sampling parameters for reasoning models.
Reference-asset generation (temporal VQA question-answer pairs) also uses temperature $0.7$; outputs are cached so that downstream evaluation remains deterministic.

\paragraph{Frontier models.}
Gemini~4~Argon, Gemini~4~Hydrogen, and Claude~Opus~5.5 are run with \emph{medium} thinking effort on the same frozen evaluation subsets, with the same task prompts, parsing, metrics, and judge configuration as the five main models, and are reported as mean and standard deviation over three runs.
The only prompt adaptation is the pixel-coordinate format for Claude~Opus~5.5 described above.
Their full results are included in Tables~\ref{tab:hoi_full}, \ref{tab:relation_extraction_full}, and~\ref{tab:visual_qa_per_category}, and their token usage in Table~\ref{tab:resources}; the size-stratified grounding, cross-judge, cross-task, and per-machine-type analyses cover the five main models only.

\paragraph{Judge model.}
All G-Eval evaluations use Gemini~2.5~Flash as the judge via the DeepEval framework~\citep{liu2023geval}.
Each criterion is scored on a $0$--$10$ rubric and linearly rescaled to $[0, 1]$ for reporting.

\subsection{Object-focused tasks}

\subsubsection{Phrase grounding}
\label{app:grounding}

\paragraph{Examples.}
Candidate frames are restricted to those (a)~that fall inside an annotated activity segment and (b)~that carry at least one annotated relation.
The surviving frames are evenly subsampled with a sampling fraction of $0.1$, from which $1{,}000$ examples are shuffled and frozen with a fixed seed.
For each frame, phrases are generated per category by a greedy selector that scores candidate descriptions by richness, preferring category together with color attributes together with a spatial relation over bare category names.
Relations derived from the annotations are expressed through templates such as ``\textit{the $\langle$color$\rangle$ $\langle$category$\rangle$ $\langle$relation$\rangle$ the $\langle$target$\rangle$}''.
At most one phrase per category is retained per frame; objects with bounding-box area below $100$~pixels are excluded.

\paragraph{System prompt.}

\begin{quote}\footnotesize\ttfamily
You are an expert at analyzing video frames to locate specific objects.\\[2pt]
For each object matching the description, output its bounding box.\\[2pt]
Bounding box format: \{coord\_desc\}, integers 0--\{scale\}. Order: \{coord\_order\}\\[2pt]
The final output should be a JSON object:\\[2pt]
\{"bboxes": [[a, b, c, d], [a, b, c, d]]\}\\[2pt]
Each inner list is exactly 4 integers. If no matching objects, return: \{"bboxes": []\}
\end{quote}

\paragraph{User prompt.}
The user prompt attaches the frame and the generated phrase:

\begin{quote}\footnotesize\ttfamily
Locate all instances of: "\{phrase\}"
\end{quote}

\paragraph{Metrics.}
We report mAP and AR averaged over IoU thresholds $[0.5{:}0.05{:}0.95]$, mAP@$0.5$, mAP@$0.75$, and size-stratified mAP/AR using the COCO size bins~\citep{lin2014coco}: small ($<{32^2}$~px), medium ($32^2$--$96^2$~px), and large ($>{96^2}$~px).

\paragraph{Full results.}
Table \ref{tab:grounding_full} reports the complete set of COCO-style detection metrics stratified by object size.
Performance drops substantially on medium-sized objects across all models, confirming that object scale remains a key bottleneck for phrase grounding in BARISTA.

\begin{table}[h]
    \centering
    \caption{\textbf{Phrase grounding (full results)} on $1000$ frozen examples. mAP/AR over IoU $\in [0.5, 0.95]$ and at IoU $0.5$, $0.75$; size bins follow COCO conventions.}
    \label{tab:grounding_full}
    \setlength{\tabcolsep}{4pt}
    \small
    \resizebox{\linewidth}{!}{%
    \begin{tabular}{@{}l cc cc cc cc c c@{}}
    \toprule
     & \multicolumn{2}{c}{All @[.5:.95]} & \multicolumn{2}{c}{Small} & \multicolumn{2}{c}{Medium} & \multicolumn{2}{c}{Large} & mAP@50 & mAP@75 \\
    \cmidrule(lr){2-3} \cmidrule(lr){4-5} \cmidrule(lr){6-7} \cmidrule(lr){8-9}
    Model & mAP & mAR & mAP & mAR & mAP & mAR & mAP & mAR & All & All \\
    \midrule
    SAM 3
        & $.362$ & $.473$
        & $.084$ & $.103$
        & $.192$ & $.245$
        & $.366$ & $.485$
        & $.393$ & $.377$ \\
    \midrule
    Gemini~3~Flash
        & $\mathbf{.534} \pm .008$ & $\mathbf{.646} \pm .002$
        & $.012 \pm .001$ & $.020 \pm .004$
        & $\mathbf{.237} \pm .019$ & $\mathbf{.292} \pm .028$
        & $\mathbf{.567} \pm .011$ & $\mathbf{.691} \pm .006$
        & $\mathbf{.697} \pm .017$ & $\mathbf{.564} \pm .007$ \\
    GPT-5.4~mini
        & $.068 \pm .014$ & $.114 \pm .017$
        & $.000 \pm .000$ & $.000 \pm .000$
        & $.001 \pm .000$ & $.003 \pm .002$
        & $.073 \pm .015$ & $.124 \pm .018$
        & $.163 \pm .022$ & $.048 \pm .011$ \\
    \midrule
    Gemma 4-31B
        & $.432 \pm .004$ & $.555 \pm .005$
        & $.002 \pm .000$ & $.005 \pm .001$
        & $.109 \pm .015$ & $.147 \pm .019$
        & $.469 \pm .002$ & $.605 \pm .003$
        & $.634 \pm .005$ & $.473 \pm .010$ \\
    Qwen3.5-27B
        & $.518 \pm .005$ & $.624 \pm .005$
        & $\mathbf{.021} \pm .021$ & $\mathbf{.026} \pm .022$
        & $.193 \pm .036$ & $.238 \pm .030$
        & $.549 \pm .007$ & $.664 \pm .003$
        & $.680 \pm .016$ & $.552 \pm .022$ \\
    \midrule
    Qwen3.5-9B
        & $.350 \pm .020$ & $.454 \pm .017$
        & $.001 \pm .001$ & $.002 \pm .000$
        & $.040 \pm .017$ & $.056 \pm .015$
        & $.393 \pm .021$ & $.507 \pm .018$
        & $.559 \pm .027$ & $.374 \pm .037$ \\
    \bottomrule
    \end{tabular}%
    }
\end{table}

\subsubsection{Egocentric hand-object interaction}
\label{app:hoi}

\paragraph{Examples.}
Interactions are extracted from the annotated \emph{human\_actions} relations.
The hand type (left, right, or both) is inferred from the source category; if the same target object appears simultaneously as the target of both a left-hand and a right-hand relation in the same frame, the two interactions are merged into a single \emph{both}-hand interaction.
The frozen evaluation set contains $1{,}000$ frames, each with at least one interaction.

\paragraph{System prompt.}
The system prompt includes three inline exemplars demonstrating the right-hand, both-hands, and two-separate-interactions cases.

\begin{quote}\footnotesize\ttfamily
You are analyzing egocentric (first-person) video frames. Identify every hand-object interaction: one or both of the camera wearer's hands actively manipulating, holding, or touching an object.\\[2pt]
For each interaction output:\\
- "hand\_boxes": a list of 1 or 2 bounding boxes --- one per interacting hand.\\
- "object\_box": a single bounding box for the object being interacted with.\\
- "hand\_type": which hand(s) are involved --- "left", "right", or "both".\\[2pt]
Each interaction is a separate element. If the same hand touches multiple objects, report each interaction separately. Bounding box format: \{coord\_desc\}, integers 0--\{scale\}.\\[2pt]
Return ONLY a JSON array --- one element per interaction. If none, return [].
\end{quote}

\paragraph{User prompt.}
The user prompt attaches the frame without additional text.

\paragraph{Metrics.}
Predicted and ground-truth interactions are matched greedily one-to-one by object-box IoU at a threshold of $0.5$, with highest-overlap predictions matched first.
On matched pairs, hand boxes are themselves matched by a secondary greedy IoU assignment.
We report interaction-level precision and recall, the mean IoU of matched hand and object boxes, the accuracy of the hand-type label on matched pairs, and a \emph{no-detection rate} computed as the fraction of frames that contain at least one ground-truth interaction but for which the model emits no prediction.
CaRe-Ego is evaluated on the same frozen evaluation set using its released inference pipeline.

\paragraph{Full results.}
Table \ref{tab:hoi_full} reports the complete set of HOI metrics, including hand-type classification accuracy.

\begin{table}[h]
    \centering
    \caption{\textbf{Egocentric HOI (full results)} on $1000$ frozen frames. Matching: greedy by object-box IoU $>0.5$. No-det.~= No-detection rate ($\downarrow$).}
    \label{tab:hoi_full}
    \setlength{\tabcolsep}{4pt}
    \small
    \resizebox{\linewidth}{!}{%
    \begin{tabular}{@{}l cccccc@{}}
        \toprule
        Model & Int. Prec. & Int. Rec. & Hand IoU & Obj. IoU & Hand-type Acc. & No-det. Rate \\
        \midrule
        CaRe-Ego & $.444$ & $.547$ & $.703$ & $.839$ & $.807$ & $.071$ \\
        \midrule
        Gemini~3~Flash
            & $.791 \pm .003$
            & $.711 \pm .011$
            & $.744 \pm .007$
            & $.892 \pm .001$
            & $.901 \pm .005$
            & $.028 \pm .007$ \\
        GPT-5.4~mini
            & $.203 \pm .015$
            & $.184 \pm .014$
            & $.483 \pm .007$
            & $.644 \pm .008$
            & $.833 \pm .006$
            & $.006 \pm .001$ \\
        \midrule
        Gemini~4~Argon
            & $\mathbf{.862} \pm .004$
            & $.777 \pm .002$
            & $\mathbf{.802} \pm .004$
            & $.894 \pm .002$
            & $.933 \pm .006$
            & $.017 \pm .002$ \\
        Gemini~4~Hydrogen
            & $.859 \pm .002$
            & $\mathbf{.788} \pm .002$
            & $.738 \pm .002$
            & $\mathbf{.898} \pm .001$
            & $\mathbf{.940} \pm .001$
            & $.009 \pm .001$ \\
        Claude~Opus~5.5
            & $.831 \pm .003$
            & $.782 \pm .002$
            & $.747 \pm .002$
            & $.884 \pm .001$
            & $.889 \pm .006$
            & $\mathbf{.004} \pm .000$ \\
        \midrule
        Gemma 4-31B
            & $.799 \pm .003$
            & $.721 \pm .002$
            & $.707 \pm .001$
            & $.857 \pm .001$
            & $.837 \pm .003$
            & $.015 \pm .002$ \\
        Qwen3.5-27B
            & $.779 \pm .008$
            & $.729 \pm .006$
            & $.706 \pm .004$
            & $.864 \pm .002$
            & $.917 \pm .005$
            & $\mathbf{.004} \pm .001$ \\
        \midrule
        Qwen3.5-9B
            & $.715 \pm .006$
            & $.583 \pm .009$
            & $.640 \pm .003$
            & $.787 \pm .005$
            & $.905 \pm .003$
            & $.121 \pm .008$ \\
        \bottomrule
    \end{tabular}%
    }
\end{table}

\subsubsection{Referring / region description}
\label{app:referring}

\paragraph{Examples.}
References are assembled deterministically from the verified annotations.
For each selected object, the reference includes:
(1)~the object category;
(2)~its attributes as key-value pairs (color, type, state);
(3)~its direct relations (e.g.\ ``coffee cup under coffee machine'', ``right hand holds coffee cup'');
and (4)~a context block for each related object, including that object's own attributes and relations.
Candidate frames are restricted to those inside annotated activity segments and subsampled with a stride of $60$ frames.
For each candidate frame, up to $10$ objects are selected from those that possess both attributes and at least one spatial or functional relation; at most $50$ examples per object category are retained.
The frozen evaluation set contains $1{,}000$ examples, each consisting of a single frame paired with a bounding box.

\paragraph{System prompt.}

\begin{quote}\footnotesize\ttfamily
You describe objects in video frames of an egocentric coffee-making scenario. You will be given a bounding box in \{coord\_desc\} format, with integer coordinates normalized to 0--\{scale\}.\\[2pt]
Write one natural sentence that:\\
1. Names the object using the most specific category you can determine.\\
2. Includes key visual attributes you can observe.\\
3. States the most important spatial or functional relation to another object. Avoid incidental proximity (near, next to, beside).\\
4. Does not mention background, surfaces, walls, or scenery.\\[2pt]
Start your final response with ``Description:'' followed by your sentence.
\end{quote}

\paragraph{User prompt.}
The user prompt attaches the frame and the query: ``Describe the object inside bounding box [y1, x1, y2, x2]''.

\paragraph{G-Eval rubric.}
Referring outputs are scored on a single \emph{Correctness} criterion ($0$--$10$ scale, rescaled to $[0,1]$).
The evaluation prompt sent to the judge includes the domain vocabulary listing all canonical category names with short descriptions, enabling the judge to assess category specificity.
The full criteria text provided to the judge is:

\begin{quote}\footnotesize
Evaluate whether the predicted referring expression correctly and specifically identifies the object in an egocentric coffee-making video frame. The reference is a structured annotation in the format:\\
\hspace*{1em}Object: $\langle$category$\rangle$\\
\hspace*{1em}Attributes: $\langle$key=value, \ldots$\rangle$\\
\hspace*{1em}Relations: $\langle$direct relations of the object$\rangle$\\
\hspace*{1em}Related objects: $\langle$attributes and relations of related objects---supplementary context$\rangle$\\[4pt]
The ``Relations'' line lists all direct relations, but the prediction is a single sentence that is not expected to mention every one. The ``Related objects'' section is supplementary context for disambiguation.\\[4pt]
A good prediction:\\
(1) Names the object with the correct or acceptably synonymous category---accept reasonable synonyms (e.g.\ ``espresso machine'' for ``portafilter machine''), but penalize significantly less specific names (e.g.\ ``machine'' for ``capsule machine'').\\
(2) States key visual attributes of the main object correctly (color, state).\\
(3) Includes at least one primary spatial or functional relation to another object (e.g.\ ``held by a hand'', ``under a capsule machine'', ``attached to a portafilter machine'').\\
(4) Is appropriately concise---a single focused sentence without scene description, background, bounding-box coordinates, or brand names.\\[4pt]
Extra correct details, equivalent wording, and minor rephrasing should never reduce the score.\\[4pt]
Domain vocabulary (canonical category names; accept reasonable synonyms, but prefer more specific names over generic ones): [\emph{full list of category names with descriptions omitted for brevity}]
\end{quote}
\begin{quote}\footnotesize
\textbf{Correctness.}\\
Score 0--3: Prediction names a completely different object category that contradicts the reference, or is unintelligible/empty.\\
Score 4--7: Correct object but with notable errors: overly generic category (e.g.\ ``coffee machine'' instead of ``capsule machine''), wrong attribute, wrong relation, omits all relations, or excessively verbose.\\
Score 8--10: Correctly identifies the object with its key attributes and at least one correct primary spatial/functional relation in a concise single sentence. Equivalent wording and extra correct details are acceptable.
\end{quote}

\subsection{Semantic tasks}

\subsubsection{Activity recognition}
\label{app:activity}

\paragraph{Examples.}
Each example consists of $F{=}4$ frames sampled uniformly from a single annotated activity segment.
Segments shorter than $F$ frames are discarded; at most $20$ examples per activity class are retained to enforce category balance.
$500$ examples are sampled across videos with a fixed seed.

\paragraph{Distractor construction.}
Each question presents $K{=}25$ choices (one correct, $24$ distractors).
When available, at least $4$ distractors are guaranteed to share the same verb as the correct answer (e.g.\ \emph{close bottle}, \emph{close cabinet}, \emph{close coffee machine}), forcing the model to discriminate based on the manipulated object rather than the action alone.
The remaining distractors are drawn randomly from the full activity vocabulary; all $25$ choices are then shuffled with a deterministic per-segment seed.

\paragraph{System prompt.}

\begin{quote}\footnotesize\ttfamily
You are an expert at fine-grained activity recognition from video frames. Focus on hand-object interactions and visible state changes across consecutive frames (e.g., direction of motion, whether an object is being placed or removed). You may provide a brief rationale, but the final line must be exactly ANSWER: <CHOICE\_ID>.
\end{quote}

\paragraph{User prompt.}
The user prompt attaches the four frames in order and poses the question with the $25$ activity labels as choices A through Y, each rendered from the display name of the corresponding activity class:

\begin{quote}\footnotesize\ttfamily
The 4 frames below are sampled uniformly from an activity segment (Frame 1 earliest, Frame 4 latest). Which activity label best describes the segment?\\[2pt]
Choices:\\
A. ...\\
B. ...\\
...
\end{quote}

\paragraph{Metrics.}
Responses are parsed by extracting the choice identifier from an ``ANSWER: \textless ID\textgreater'' line (case-insensitive, allowing optional rationale before the final line).
We report top-$1$ accuracy over all examples (parse failures scored as incorrect).

\subsubsection{Relation extraction}
\label{app:relation_extraction}

\paragraph{Examples.}
Candidate frames are subsampled with a temporal stride of $300$ frames and filtered to those carrying at least one annotated relation.
$1{,}000$ frames are shuffled and frozen with a fixed seed.

\paragraph{Set-of-mark overlay.}
Each frame is shown twice: once as the original RGB frame and once as a \emph{set-of-mark} overlay in which the instance masks of annotated objects are blended with the frame at alpha $0.5$, their boundaries are drawn at alpha $0.9$, and each instance is labeled with an integer identifier at font size~$20$.
A legend mapping identifier to category is included in the user prompt.
The allowed relation vocabulary is derived from all video annotations and appended to the system prompt, listing for each relation type the valid set of value strings.

\paragraph{System prompt.}

\begin{quote}\footnotesize\ttfamily
You extract pairwise relations between objects in a coffee-making scene. Each object has a numeric ID on the set-of-mark image and in the ID-to-category list.\\[2pt]
The final response must be a JSON array of relation objects. If there are no relations, return []. Each element must be an object with keys: source\_id (integer), target\_id (integer), relation\_type (string), value (string). Use only relation\_type and value strings that appear together under the same relation\_type in the user message.\\[2pt]
Example: [\{"source\_id": 1, "target\_id": 2, "relation\_type": "position", "value": "part of"\}]
\end{quote}

\paragraph{User prompt.}
The user prompt attaches the original frame, the set-of-mark overlay, and the identifier-to-category mapping.

\paragraph{Metrics.}
Predicted and ground-truth relations are compared as multisets of tuples $(s, t, r, v)$ after case- and whitespace-normalization of all string fields.
We report micro-averaged precision, recall, and F1 pooled over all frames, together with the same metrics broken down by relation type.

\paragraph{Full results.}
Table \ref{tab:relation_extraction_full} reports the complete precision, recall, and F1 breakdown.
Among the five main models, Qwen3.5-27B obtains the highest recall (.799), while Gemma~4-31B achieves the highest precision (.736); the frontier models improve both, up to .876 recall (Gemini~4~Argon) and .778 precision (Claude~Opus~5.5).

\begin{table}[ht]
    \centering
    \caption{\textbf{Relation extraction (full results)}. Micro P/R/F1 under exact multiset matching; per-type micro F1.}
    \label{tab:relation_extraction_full}
    \setlength{\tabcolsep}{4pt}
    \small
    \begin{tabular}{@{}l ccc cc@{}}
        \toprule
        & \multicolumn{3}{c}{Overall} & \multicolumn{2}{c}{Per-type micro F1} \\
        \cmidrule(lr){2-4} \cmidrule(lr){5-6}
        Model 
        & $P_{\mathrm{micro}}$ 
        & $R_{\mathrm{micro}}$ 
        & $\mathrm{F1}_{\mathrm{micro}}$ 
        & \texttt{human\_actions} 
        & \texttt{position} \\
        \midrule
        Gemini~3~Flash
            & $.655 \pm .006$
            & $.697 \pm .009$
            & $.675 \pm .008$
            & $.675 \pm .004$
            & $.710 \pm .007$ \\
        GPT-5.4~mini
            & $.546 \pm .009$
            & $.586 \pm .008$
            & $.565 \pm .008$
            & $.534 \pm .011$
            & $.569 \pm .008$ \\
        \midrule
        Gemini~4~Argon
            & $.755 \pm .002$
            & $\mathbf{.876} \pm .002$
            & $.811 \pm .002$
            & $.698 \pm .005$
            & $.819 \pm .002$ \\
        Gemini~4~Hydrogen
            & $.747 \pm .002$
            & $.857 \pm .001$
            & $.798 \pm .002$
            & $\mathbf{.742} \pm .003$
            & $.802 \pm .002$ \\
        Claude~Opus~5.5
            & $\mathbf{.778} \pm .001$
            & $.874 \pm .001$
            & $\mathbf{.823} \pm .000$
            & $.730 \pm .006$
            & $\mathbf{.829} \pm .001$ \\
        \midrule
        Gemma 4-31B
            & $.736 \pm .002$
            & $.767 \pm .002$
            & $.751 \pm .002$
            & $.673 \pm .012$
            & $.758 \pm .001$ \\
        Qwen3.5-27B
            & $.699 \pm .006$
            & $.799 \pm .003$
            & $.745 \pm .005$
            & $.626 \pm .006$
            & $.756 \pm .004$ \\
        \midrule
        Qwen3.5-9B
            & $.587 \pm .005$
            & $.642 \pm .002$
            & $.613 \pm .004$
            & $.573 \pm .010$
            & $.630 \pm .005$ \\
        \bottomrule
    \end{tabular}
\end{table}

\subsubsection{Temporal visual question answering}
\label{app:visual_qa}

\paragraph{Examples.}
Reference QA pairs are produced offline by Gemini~2.5~Flash Lite at temperature~$0.7$.
For every candidate clip the generator detects annotation diffs between the clip's start and end windows---which object attributes changed, which hand-contact sequences occurred, which spatial relations shifted, and which activity labels transitioned---and instantiates one QA pair per non-empty category.
Clip starts are strided every $144$ frame indices (non-overlapping); each clip contains $6$ keyframes spaced $24$ frame indices apart, spanning four seconds at $30$~fps.
The generator's system prompt is:

\begin{quote}\footnotesize\ttfamily
You are creating category-tagged question--answer pairs for a visual QA benchmark about coffee-making videos captured from an egocentric viewpoint. [...] Generate exactly 1 question--answer pair for the assigned category. The question should naturally encompass all listed changes of that type. [...] Return ONLY a valid JSON object: \{"question": "...", "answer": "...", "category": "..."\}
\end{quote}

QA pairs are cached to disk and reused verbatim across every subsequently evaluated model.
$300$ pairs are sampled uniformly across the four categories (\emph{state change}, \emph{action transition}, \emph{relation change}, \emph{activity progression}; up to $100$ per category) to form the frozen evaluation set.

\paragraph{System prompt.}

\begin{quote}\footnotesize\ttfamily
You are an expert at visual question answering for coffee-making videos. You will receive a sequence of ordered frames sampled from a video clip, along with a question about the content. Analyze the frames carefully and provide a detailed, accurate answer. Your answer should be specific, clear, and based solely on what is visible in the frames. Start your final response with "Answer:" followed by your answer.
\end{quote}

\paragraph{User prompt.}
The user prompt attaches the frames (up to six) in order and poses the reference question verbatim.
Answers are extracted by taking the text after the final ``Answer:'' marker (case-insensitive).

\paragraph{G-Eval rubrics.}
Both temporal VQA criteria are scored on the $0$--$10$ scale defined below and are rescaled to $[0,1]$ for reporting.

\begin{quote}\footnotesize
\textbf{Accuracy.}  Are all claims in the prediction factually correct per the reference? Only penalize claims that are directly contradicted---wrong change direction, wrong object name, or wrong contact state. Do not penalize omissions, extra details, or equivalent wording. Contact-action synonyms (touching/holding/gripping the same object) are the same state. Machine synonyms (coffee machine / espresso machine / capsule machine) are always acceptable. For action-transition questions, evaluate whether the contact targets in the prediction are correct---do not penalize for omitting or including temporal anchors like ``at the start'' or ``by the end'', since the hand may appear at any point in the clip.\\[4pt]
Score 0--3: The prediction contains clearly wrong claims that directly contradict the reference---reversed change directions, wrong object names, or a stated contact state that is inconsistent with the reference.\\
Score 4--7: The prediction is mostly correct but includes at least one claim that directly contradicts the reference (e.g.\ wrong change direction or wrong object name). Additional details not in the reference should not place the answer here.\\
Score 8--10: All key claims in the prediction are correct and consistent with the reference. Additional correct details, plausible intermediate steps, or non-contact observations not in the reference are acceptable. Equivalent terminology is acceptable.
\end{quote}
\begin{quote}\footnotesize
\textbf{Completeness.}  Does the prediction cover every change listed in the reference? Do not penalize incorrect claims---only missing ones. Idle hand states between contact events do not need to be stated explicitly. Per-hand attribution (left vs.\ right) is not required. Semantically equivalent descriptions are acceptable. For action-transition questions, evaluate whether every contact target in the reference is mentioned in the prediction---ignore temporal anchors like ``at the start'' or ``by the end'' since the hand may appear at any point in the clip.\\[4pt]
Score 0--3: The prediction omits most changes---it gives a vague or generic response that skips the specific objects, states, or transitions from the reference.\\
Score 4--7: The prediction mentions some changes but misses at least one important object, state transition, or activity step present in the reference.\\
Score 8--10: The prediction covers all changes from the reference. Minor rephrasing is acceptable.
\end{quote}

\paragraph{Scope of the action-transition rubric.}
Both rubrics ignore temporal anchors for action-transition questions because clips span only four seconds and a hand may enter or leave contact at any point, so references and predictions legitimately differ in phrases such as ``at the start''.
As a consequence, action-transition scores primarily reflect whether the correct contact targets are identified and can reward naming the involved objects without recovering their exact order.

\paragraph{Full results.}
Table \ref{tab:visual_qa_per_category} breaks down temporal VQA performance by question category.
GPT-5.4~mini performs particularly well on state-change questions, while relation-change questions are easiest across models and action transitions are hardest.
The Gemini~4 models gain mainly on state-change and activity-progression questions; action transitions remain the hardest category for all three frontier models and do not improve over the best main model.

\begin{table}[h]
    \centering
    \caption{\textbf{Temporal VQA per-category mean} (G-Eval mean of Accuracy and Completeness, in $[0,1]$).}
    \label{tab:visual_qa_per_category}
    \setlength{\tabcolsep}{4pt}
    \small
    \begin{tabular}{@{}l cccc@{}}
        \toprule
        Model & Relation change & Action transition & State change & Activity progression \\
        \midrule
        Gemini~3~Flash   & $0.786 \pm .019$ & $0.548 \pm .012$ & $0.657 \pm .010$ & $0.603 \pm .023$ \\
        GPT-5.4~mini     & $0.760 \pm .023$ & $0.525 \pm .004$ & $0.817 \pm .015$ & $0.543 \pm .013$ \\
        \midrule
        Gemini~4~Argon & $\mathbf{0.843} \pm .010$ & $0.565 \pm .045$ & $\mathbf{0.879} \pm .014$ & $0.656 \pm .017$ \\
        Gemini~4~Hydrogen & $0.830 \pm .006$ & $0.510 \pm .001$ & $0.833 \pm .012$ & $\mathbf{0.780} \pm .022$ \\
        Claude~Opus~5.5 & $0.837 \pm .006$ & $0.395 \pm .022$ & $0.738 \pm .022$ & $0.547 \pm .005$ \\
        \midrule
        Gemma 4-31B      & $0.785 \pm .007$ & $0.470 \pm .026$ & $0.630 \pm .015$ & $0.607 \pm .023$ \\
        Qwen3.5-27B      & $0.826 \pm .010$ & $\mathbf{0.575} \pm .034$ & $0.743 \pm .009$ & $0.585 \pm .008$ \\
        \midrule
        Qwen3.5-9B       & $0.793 \pm .043$ & $0.486 \pm .019$ & $0.732 \pm .033$ & $0.509 \pm .019$ \\
        \bottomrule
    \end{tabular}
\end{table}

\subsection{Runtime and resources}
\label{app:runtime}

All experiments with open-weight models were run on the same serving setup within each model scale.
For Qwen3.5-27B and Gemma 4-31B, each run used two NVIDIA H200 GPUs with 128\,GB system RAM in total.
We served one VLM instance per GPU and used data parallelism (degree $2$); requests were processed with a total concurrency of $4$.
For Qwen3.5-9B, we used the same serving configuration, but ran the model on NVIDIA RTX PRO 6000 GPUs, as the smaller model required less memory.

\textbf{Token usage.}
We report the average number of input and output tokens per example for each model--task pair in Table \ref{tab:resources}.
Token counts reflect provider-specific tokenization and include any system-prompt overhead.
Claude~Opus~5.5 has higher input counts because images are sent at full resolution.

\begin{table}[h]
\centering
\small
\caption{
\textbf{Token usage overview.}
Average token usage per example across all benchmark tasks and models.
In = input tokens, Out = output tokens (including thinking tokens).
}
\label{tab:resources}
\resizebox{\linewidth}{!}{%
\begin{tabular}{l rr rr rr rr rr rr rr rr}
\toprule
 & \multicolumn{2}{c}{Gemini~3~Flash} & \multicolumn{2}{c}{GPT-5.4~mini} & \multicolumn{2}{c}{Qwen3.5-27B} & \multicolumn{2}{c}{Gemma 4-31B} & \multicolumn{2}{c}{Qwen3.5-9B} & \multicolumn{2}{c}{Gemini~4~Argon} & \multicolumn{2}{c}{Gemini~4~Hydrogen} & \multicolumn{2}{c}{Claude~Opus~5.5} \\
\cmidrule(lr){2-3} \cmidrule(lr){4-5} \cmidrule(lr){6-7} \cmidrule(lr){8-9} \cmidrule(lr){10-11} \cmidrule(lr){12-13} \cmidrule(lr){14-15} \cmidrule(lr){16-17}
Task 
& In & Out
& In & Out
& In & Out
& In & Out
& In & Out
& In & Out
& In & Out
& In & Out \\
\midrule
Grounding            & 1236 & 228  & 1217 & 173  & 2365 & 704   & 1272 & 425  & 2365 & 894 & 1265 & 66 & 1265 & 562 & 3152 & 105 \\
Hand-object          & 1581 & 104  & 1494 & 174  & 2626 & 1543  & 1628 & 602  & 2626 & 2026 & 1605 & 100 & 1605 & 705 & 3414 & 187 \\
Referring            & 1254 & 385  & 1214 & 165  & 2424 & 1443  & 1288 & 433  & 1615 & 1148 & 1287 & 37 & 1293 & 511 & 3208 & 215 \\
Relation extraction  & 2521 & 524  & 2547 & 332  & 4522 & 3829  & 2573 & 1134 & 4526 & 4911 & 2559 & 226 & 2559 & 952 & 5975 & 321 \\
Activity recognition & 4617 & 559  & 4604 & 185  & 9267 & 1911  & 4661 & 945  & 9267 & 2473 & 4659 & 115 & 4659 & 621 & 12327 & 203 \\
Temporal VQA         & 6585 & 304  & 6809 & 136  & 12034 & 1792 & 6643 & 545  & 12058 & 2254 & 6641 & 108 & 6641 & 582 & 15982 & 343 \\
\bottomrule
\end{tabular}%
}
\end{table}
\newpage

\section{BARISTA's characteristics}
\label{app:dataset_stats}

\subsection{Distributions}
\label{app:dataset_distributions}

\begin{figure}[h]
  \centering
  \includegraphics[width=\linewidth]{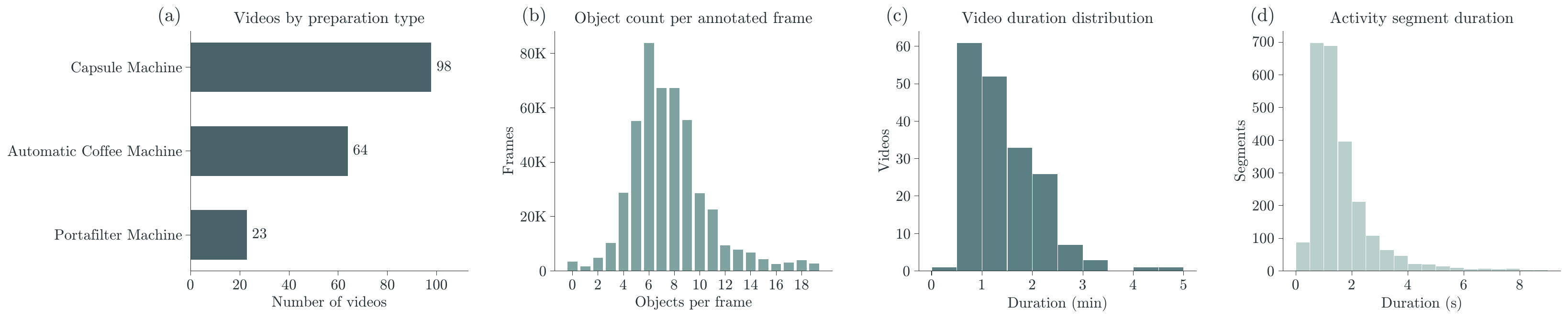}
  \caption{\textbf{Dataset overview.}
    (a)~Video distribution across preparation types.
    (b)~Number of annotated objects per frame.
    (c)~Video duration distribution.
    (d)~Activity segment duration.}
  \label{fig:dataset_overview}
\end{figure}

Figure \ref{fig:dataset_overview} presents key distributional properties of BARISTA.
Capsule machines account for the majority of recordings, followed by fully automatic and portafilter machines~(a).
Annotated frames typically contain 6--10 simultaneously tracked objects~(b), reflecting the dense, multi-object nature of coffee preparation scenes.
Most videos fall in the 1--3\,min range~(c), while activity segment durations range from sub-second manipulations to under one minute~(d).

\begin{figure}[h]
  \centering
  \includegraphics[width=\linewidth]{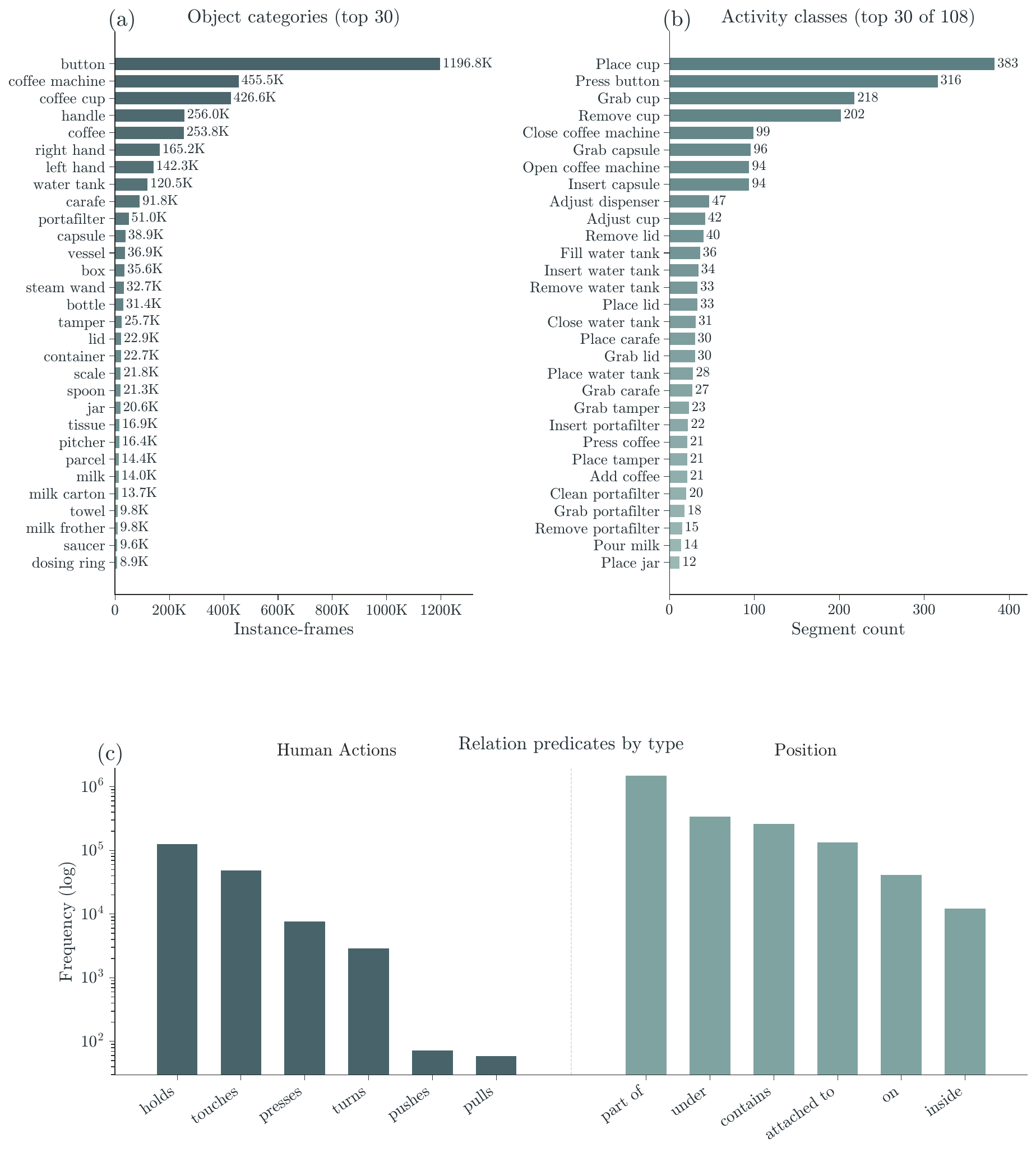}
  \caption{\textbf{Annotation distributions.}
    (a)~Object category frequency (top 30 of 46 categories).
    (b)~Activity class frequency (top 30 of 108 classes).
    (c)~Relation predicate frequencies by type (log scale).}
  \label{fig:distributions}
\end{figure}

Figure \ref{fig:distributions} shows the annotation distributions.
Object category counts~(a) exhibit a long tail: high-frequency categories such as \emph{button}, \emph{cup}, and \emph{coffee machine} appear in nearly every video, while specialized items occur only in specific preparation styles.
Activity classes~(b) are similarly skewed: common actions such as \emph{Place cup} and \emph{Press button} occur hundreds of times.
Relation predicates~(c) are grouped by type: hand-object interactions (\emph{holds}, \emph{touches}) dominate the \emph{human actions} group, while spatial predicates (\emph{part of}, \emph{under}, \emph{contains}) capture the static scene structure.

\subsection{Recording diversity}
\label{app:diversity}

Table~\ref{tab:diversity} summarizes the diversity of the recordings, computed from the released metadata ($185$ videos, $4.38$\,h, $472{,}764$ frames, of which $469{,}201$ are annotated).

\begin{table}[!htb]
\centering
\small
\caption{\textbf{Recording diversity of BARISTA.} Counts are numbers of videos unless stated otherwise.}
\label{tab:diversity}
\setlength{\tabcolsep}{6pt}
\begin{tabular}{@{}l p{0.68\linewidth}@{}}
\toprule
Axis & Values \\
\midrule
Preparation type & capsule $98$, fully automatic $64$, portafilter $23$ \\
Recording device & iPhone (12~mini / 13~Pro / 15 / 15~Pro / 16~Pro) $72$, Ray-Ban Meta (Stories, Wayfarer) $63$, Apple Vision Pro $35$, Samsung S24~Ultra $10$, DigiLens ARGO $5$ \\
Resolution & $20$ distinct resolutions, portrait and landscape, from $720{\times}1280$ to $1920{\times}1080$ \\
Unique recorders & $10$ \\
Distinct kitchens / locations & $14$ \\
Distinct coffee machines & $15$ \\
\bottomrule
\end{tabular}
\end{table}

\subsection{Annotated examples and data access}
\label{app:annotation_examples}

\begin{figure}[h!]
  \centering
  \includegraphics[width=\linewidth]{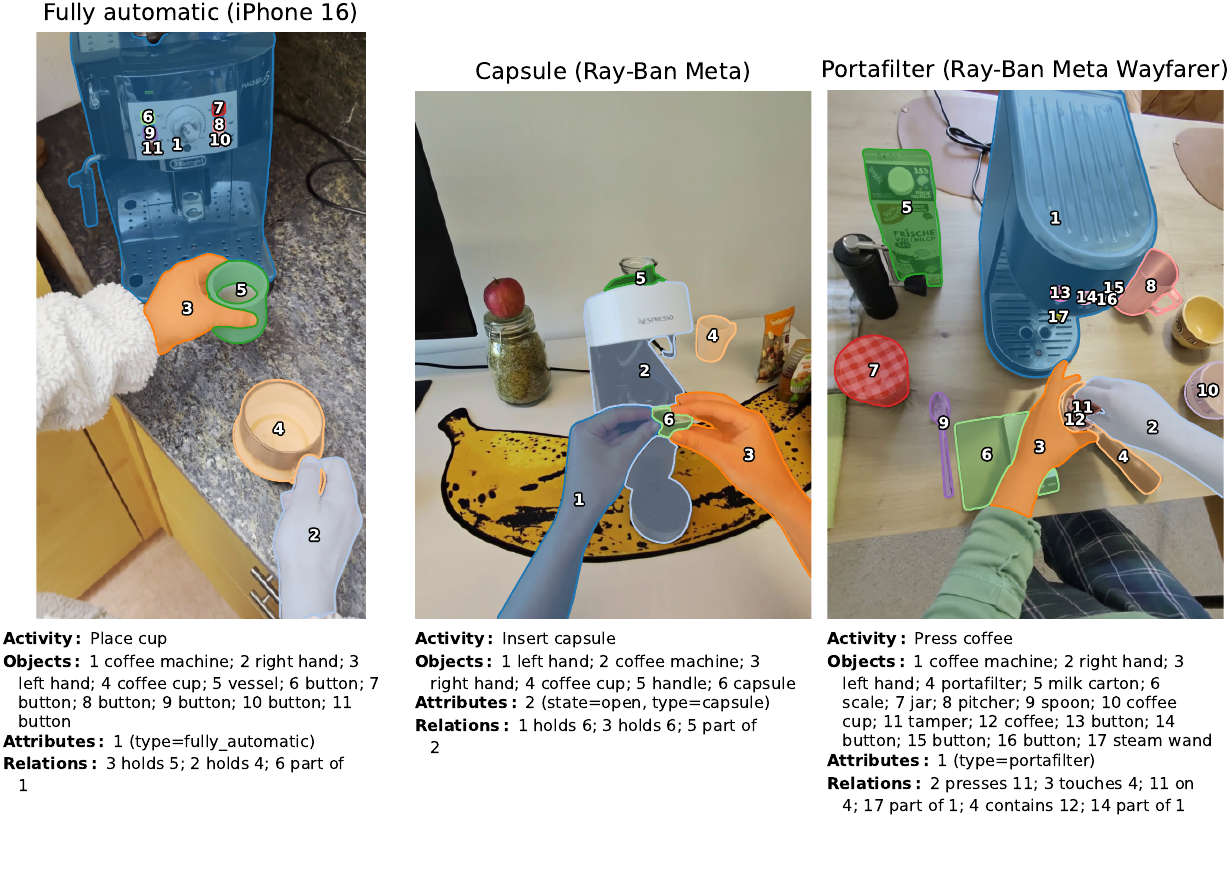}
  \caption{\textbf{Annotated example frames.} One frame per preparation style (recording device in parentheses) with its instance masks; numbers index object instances. Below each frame: the activity label, the object categories, selected attributes, and selected typed relations (\emph{source} \emph{predicate} \emph{target}) from the same per-frame scene graph.}
  \label{fig:annotation_examples}
\end{figure}

The complete dataset, including all raw videos and the per-frame annotations (masks, boxes, attributes, relations, hand-object interactions, activities, and process steps), is available through the anonymized repository linked in footnote~1 on the first page.
So that the annotations can be inspected without downloading the dataset, Figure~\ref{fig:annotation_examples} shows one annotated frame per preparation style together with the co-registered object, attribute, relation, and activity annotations of that frame.

\section{Matched diagnostic protocol}
  \label{app:failure_analysis}

Here, we describe the matched diagnostic cases used in Section~\ref{sec:failure_analysis}.
The purpose of these cases is to compare several deterministic tasks on the same video context.

\subsection{Matched cases and co-failure analysis}
\label{app:matched_cases}

\paragraph{Case construction.}
We construct $200$ cases from activity segments that have an immediately following annotated activity in the same video.
Each case receives one shared identifier.
Grounding, hand-object interaction, and relation extraction use the same anchor frame.
Activity recognition uses four frames from the surrounding activity segment.
For transition recognition, we sample eight ordered frames from the start of the current activity to the end of the following activity.
The model selects the annotated transition between these two activities from multiple choices.

This transition-recognition task is different from the benchmark's open-ended temporal VQA task.
Temporal VQA asks free-form questions about state, action, relation, and activity changes, and is evaluated with an LLM judge.
The diagnostic task instead uses a fixed multiple-choice answer and exact accuracy, which allows a deterministic pass/fail record for every case.

\paragraph{Scoring.}
A grounding prediction passes when its mean best IoU is at least $0.5$.
A hand-object prediction passes when matched interaction recall is at least $0.5$.
A relation prediction passes when example-level relation F1 is at least $0.5$.
Activity and transition predictions pass when the selected answer is correct.
Malformed or missing outputs are marked as unscored.
Stage pass rates are computed only over scored outputs, while the ``Valid'' column in
Table~\ref{tab:failure_analysis} reports the fraction of all stage--case outputs that received a valid score.

\paragraph{Case-level records.}
For every model, the released diagnostic record stores one pass, fail, or unscored value for each of the five tasks and each shared case identifier.
This makes it possible to inspect a single video and identify the tasks on which a model was unreliable.
The five tasks are evaluated with separate prompts; therefore, the records describe an error profile for the same video rather than a sequence of errors produced by one model computation.

\paragraph{Conditional co-failure analysis.}
For each ordered pair of diagnostic tasks, we compare the probability that a target task fails when a source task fails with the probability that the same target task fails when the source task passes.
Both outcomes must be valid for a case to enter the comparison.
We report the difference between these conditional failure rates and compute $95\%$ confidence intervals by resampling videos.

Qwen3.5-9B shows the clearest association with activity recognition: activity failure is $17.0$ percentage points more likely after a grounding failure ($27.9\%$ versus $11.0\%$; $95\%$ CI $[3.5,30.6]$) and $17.5$ points more likely after a relation-extraction failure ($29.5\%$ versus
  $12.1\%$; $[4.8,30.3]$).
The corresponding intervals for Qwen3.5-27B and Gemma~4-31B include zero.
No source task has a reliable association with transition-recognition failure, which is rare for all models.

\subsection{Evidence-handoff intervention}
\label{app:handoff}

To test whether structured perceptual evidence helps process-level reasoning, we repeat activity and transition recognition on the $200$ matched cases, adding a JSON block with anchor-frame grounding phrases and boxes, hand-object interactions and boxes, and relation tuples.
We provide either verified BARISTA annotations or the model's own predictions, explicitly identifying their source.
The model uses both the ordered frames and this evidence, resolving conflicts from the images.
We compare each condition with the same model's image-only baseline on valid paired cases ($181$--$200$ per cell), reporting accuracy changes with $95\%$ video-clustered bootstrap confidence intervals ($2{,}000$ resamples).

Verified evidence significantly improves activity recognition only for Gemma~4-31B ($72.0\% \rightarrow 78.5\%$), which has the weakest grounding in Table~\ref{tab:failure_analysis}; its own predictions provide a similar gain (Table~\ref{tab:handoff}).
Neither Qwen model shows a significant change, nor does any model on transition recognition, where accuracy is already near ceiling.
The benefit therefore depends on the model and task, supporting separate evaluation of perceptual and process-level capabilities.

\begin{table}[!htb]
\centering
\small
\caption{\textbf{Evidence handoff on the $200$ matched cases.} Change in accuracy (pp) relative to the image-only baseline when ground-truth (GT) or predicted anchor-frame evidence is added to the prompt, with video-clustered bootstrap $95\%$ CIs. Bold: CI excludes zero.}
\label{tab:handoff}
\setlength{\tabcolsep}{5pt}
\begin{tabular}{@{}l cc cc@{}}
\toprule
 & \multicolumn{2}{c}{Activity recognition} & \multicolumn{2}{c}{Transition recognition} \\
\cmidrule(lr){2-3} \cmidrule(lr){4-5}
Model & $\Delta$ GT evidence & $\Delta$ predicted evidence & $\Delta$ GT evidence & $\Delta$ predicted evidence \\
\midrule
Qwen3.5-9B  & $-1.1\ [-4.8, 2.5]$ & $+1.7\ [-3.0, 6.0]$ & $-1.0\ [-3.1, 1.1]$ & $-1.0\ [-3.1, 1.0]$ \\
Qwen3.5-27B & $-1.1\ [-6.7, 4.5]$ & $\phantom{+}0.0\ [-4.3, 4.1]$ & $+0.5\ [-1.1, 2.4]$ & $+1.0\ [-0.9, 3.1]$ \\
Gemma 4-31B & $\mathbf{+6.5}\ [2.5, 10.8]$ & $\mathbf{+5.5}\ [0.9, 10.3]$ & $-1.0\ [-3.6, 1.4]$ & $-0.5\ [-1.6, 0.0]$ \\
\bottomrule
\end{tabular}
\end{table}

\section{Robustness and cross-task analyses}
\label{app:analyses}

\subsection{LLM as a judge: Cross-judge robustness}
\label{app:judge_robustness}

To assess the sensitivity of G-Eval scores to judge identity, we re-score all five models on both judge-dependent tasks---referring expression generation ($1{,}000$ examples) and temporal VQA ($300$ examples)---using GPT-5.4~mini as an alternative judge, replacing Gemini~2.5~Flash used in the main evaluation.
The two judges belong to different model families (Google vs.\ OpenAI), providing a cross-family consistency check.

\paragraph{Ranking preservation.}
Table \ref{tab:judge_referring} and Table \ref{tab:judge_vqa} report per-model scores under both judges.
We compute Kendall~$\tau$-b rank correlation across the five models for each metric.
On referring (Correctness), the model ranking is identical under both judges ($\tau = 1.0$).
On temporal VQA, the ranking is preserved for mean score ($\tau = 1.0$) and Completeness ($\tau = 1.0$); for Accuracy alone, a single adjacent swap occurs between Gemini~3~Flash and GPT-5.4~mini---models that were already within $0.004$ of each other under the primary judge---yielding $\tau = 0.8$.

\begin{table}[h]
\centering
\small
\caption{
\textbf{Referring expression} G-Eval Correctness under two judges.
Model ranking is identical ($\tau = 1.0$).
}
\label{tab:judge_referring}
\setlength{\tabcolsep}{6pt}
\begin{tabular}{@{}l cc c@{}}
\toprule
Model & Gemini 2.5 Flash & GPT-5.4~mini & $\Delta$ \\
\midrule
Gemini~3~Flash & $.757$ & $.643$ & $-.114$ \\
Gemma 4-31B    & $.669$ & $.598$ & $-.071$ \\
Qwen3.5-27B    & $.623$ & $.542$ & $-.081$ \\
Qwen3.5-9B    & $.569$ & $.519$ & $-.050$ \\
GPT-5.4~mini   & $.302$ & $.320$ & $+.018$ \\
\bottomrule
\end{tabular}
\end{table}

\begin{table}[h]
\centering
\small
\caption{
\textbf{Temporal VQA} G-Eval scores under two judges.
Ranking is preserved for Mean ($\tau = 1.0$) and Completeness ($\tau = 1.0$); Accuracy has one adjacent swap ($\tau = 0.8$).
}
\label{tab:judge_vqa}
\setlength{\tabcolsep}{4pt}
\resizebox{\linewidth}{!}{%
\begin{tabular}{@{}l cc c cc c cc c@{}}
\toprule
& \multicolumn{3}{c}{Mean} & \multicolumn{3}{c}{Accuracy} & \multicolumn{3}{c}{Completeness} \\
\cmidrule(lr){2-4} \cmidrule(lr){5-7} \cmidrule(lr){8-10}
Model & Gemini & GPT & $\Delta$ & Gemini & GPT & $\Delta$ & Gemini & GPT & $\Delta$ \\
\midrule
Qwen3.5-27B    & $.702$ & $.622$ & $-.080$ & $.727$ & $.596$ & $-.131$ & $.683$ & $.648$ & $-.035$ \\
GPT-5.4~mini   & $.683$ & $.607$ & $-.076$ & $.701$ & $.586$ & $-.114$ & $.671$ & $.627$ & $-.044$ \\
Gemini~3~Flash & $.674$ & $.599$ & $-.075$ & $.705$ & $.577$ & $-.129$ & $.650$ & $.620$ & $-.030$ \\
Qwen3.5-9B    & $.666$ & $.587$ & $-.079$ & $.691$ & $.569$ & $-.121$ & $.645$ & $.604$ & $-.041$ \\
Gemma 4-31B    & $.647$ & $.570$ & $-.076$ & $.665$ & $.552$ & $-.113$ & $.630$ & $.588$ & $-.041$ \\
\bottomrule
\end{tabular}%
}
\end{table}

\paragraph{Absolute calibration shift.}
The GPT-5.4~mini judge assigns systematically lower scores than Gemini~2.5~Flash (mean $\Delta \approx -0.07$ on referring, $-0.08$ on VQA), indicating a stricter calibration.
However, the shift is approximately uniform across models: the standard deviation of $\Delta$ across models is $0.023$ on referring and $0.002$ on VQA mean, indicating that the offset is judge-specific rather than model-specific.

\paragraph{Self-preference probe.}
Gemini~3~Flash---evaluated by its own family's judge in the primary setup---retains rank~1 on referring under the cross-family GPT-5.4~mini judge ($0.643$ vs.\ next-best $0.598$).
Conversely, GPT-5.4~mini's own referring predictions do not receive a score boost when judged by GPT-5.4~mini ($0.320$ vs.\ $0.302$ under Gemini), remaining last-ranked under both judges.
These observations provide evidence against systematic self-preference effects for either judge family on the tasks and rubrics used in BARISTA.

\paragraph{Per-category VQA analysis.}
Table \ref{tab:judge_vqa_category} reports the per-category mean scores under both judges.
The cross-judge ranking is consistent within categories: relation-change questions are easiest under both judges, and action transitions remain hardest.

\begin{table}[h]
\centering
\small
\caption{
\textbf{Temporal VQA per-category mean} under both judges.
}
\label{tab:judge_vqa_category}
\setlength{\tabcolsep}{4pt}
\resizebox{\linewidth}{!}{%
\begin{tabular}{@{}l cccc cccc@{}}
\toprule
& \multicolumn{4}{c}{Gemini 2.5 Flash judge} & \multicolumn{4}{c}{GPT-5.4~mini judge} \\
\cmidrule(lr){2-5} \cmidrule(lr){6-9}
Model & Rel.\ chg & State chg & Act.\ prog & Act.\ trans & Rel.\ chg & State chg & Act.\ prog & Act.\ trans \\
\midrule
Qwen3.5-27B    & $.823$ & $.743$ & $.584$ & $.596$ & $.786$ & $.684$ & $.510$ & $.357$ \\
GPT-5.4~mini   & $.772$ & $.804$ & $.535$ & $.522$ & $.744$ & $.735$ & $.455$ & $.327$ \\
Gemini~3~Flash & $.805$ & $.653$ & $.623$ & $.535$ & $.781$ & $.562$ & $.547$ & $.382$ \\
Qwen3.5-9B    & $.841$ & $.714$ & $.527$ & $.470$ & $.768$ & $.629$ & $.463$ & $.349$ \\
Gemma 4-31B    & $.778$ & $.638$ & $.601$ & $.468$ & $.725$ & $.544$ & $.542$ & $.336$ \\
\bottomrule
\end{tabular}%
}
\end{table}

\paragraph{Summary.}
Model rankings on both judge-dependent tasks are robust to judge identity ($\tau \geq 0.8$ across all metrics, $\tau = 1.0$ on three of four).
The single observed swap involves two models separated by $< 0.005$ under the primary judge.
These results support the validity of the G-Eval protocol used in the main evaluation and indicate that the reported rankings reflect genuine performance differences rather than judge-specific artifacts.

\subsection{Cross-task decomposability analysis}
\label{app:decomposability}

A core design claim of BARISTA is that its task families isolate complementary capabilities, so that performance on one part of the benchmark need not track performance on another.
We test this claim post-hoc on the benchmark predictions.

\paragraph{Setup.}
For every task we compute a per-example score (mean best IoU for grounding, per-frame interaction $F_1$ for HOI, tuple $F_1$ for relation extraction, $0/1$ correctness for activity MCQ, G-Eval mean for referring and temporal VQA), and pool these scores across examples and all $3$ seeds to obtain a single mean per (model, task) pair.
We then compute Kendall~$\tau$-b on the $5$-model rankings induced by every pair of tasks.
A high $\tau$ between two tasks means they order the evaluated models almost identically (and so likely measure the same capability); a low or negative $\tau$ means one task ranks models in a way the other cannot predict.

\begin{figure}[h]
    \centering
    \includegraphics[width=0.7\linewidth]{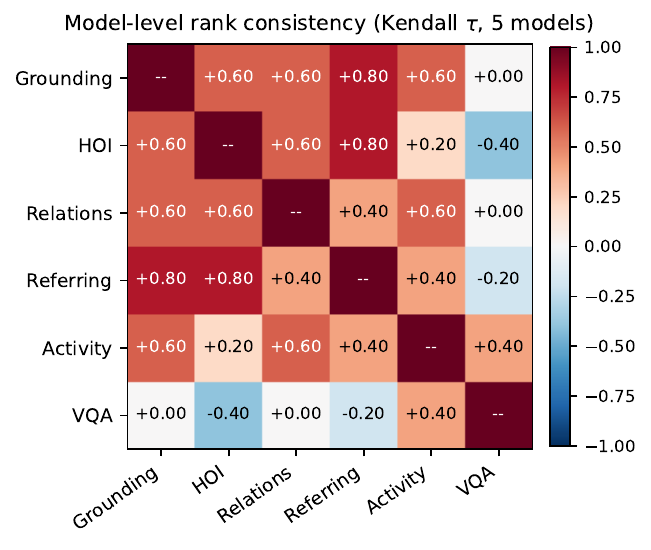}
    \caption{\textbf{Cross-task decomposability.} Kendall $\tau$-b between the rankings of the five evaluated VLMs across each task pair. The structural cluster (Grounding/HOI/Relations/Referring) is internally consistent, while Temporal VQA is decoupled from object-focused capabilities.}
    \label{fig:cross_task_kendall}
\end{figure}

\paragraph{Result.}
Table \ref{tab:decomposability_kendall} and Figure \ref{fig:cross_task_kendall} report the full $\tau$ matrix.
The four object-focused / structural tasks (grounding, HOI, relations, referring) form an internally consistent cluster ($\tau \geq 0.6$ on $5$ of $6$ pairs).
Activity recognition is moderately aligned with the structural cluster ($\tau \in [0.2, 0.6]$).
Temporal VQA, by contrast, is decoupled from grounding ($\tau{=}0.0$), HOI ($\tau{=}-0.4$), relations ($\tau{=}0.0$) and referring ($\tau{=}-0.2$); it correlates weakly only with activity recognition ($\tau{=}0.4$).
Although based on a small number of models, this rank analysis provides initial evidence that BARISTA does not collapse to a single ``video competence'' axis, but instead captures multiple distinct dimensions of video understanding.
Concretely, GPT-5.4~mini ranks last on every object-focused task yet ranks among the top two on temporal VQA, while the strongest grounding models rank lower on VQA --- a pattern that would be invisible to any single-axis benchmark.

\begin{table}[h]
\centering
\small
\caption{\textbf{Model-level rank consistency.} Kendall $\tau$-b between the rankings of the five evaluated VLMs induced by each pair of tasks (computed on per-task pooled mean scores).}
\label{tab:decomposability_kendall}
\setlength{\tabcolsep}{6pt}
\begin{tabular}{@{}l cccccc@{}}
\toprule
 & Grounding & HOI & Relations & Referring & Activity & VQA \\
\midrule
Grounding & --     & $+0.6$ & $+0.6$ & $+0.8$ & $+0.6$ & $+0.0$ \\
HOI       & $+0.6$ & --     & $+0.6$ & $+0.8$ & $+0.2$ & $-0.4$ \\
Relations & $+0.6$ & $+0.6$ & --     & $+0.4$ & $+0.6$ & $+0.0$ \\
Referring & $+0.8$ & $+0.8$ & $+0.4$ & --     & $+0.4$ & $-0.2$ \\
Activity  & $+0.6$ & $+0.2$ & $+0.6$ & $+0.4$ & --     & $+0.4$ \\
VQA       & $+0.0$ & $-0.4$ & $+0.0$ & $-0.2$ & $+0.4$ & --     \\
\bottomrule
\end{tabular}
\end{table}

\subsection{Per-machine-type breakdown}
\label{app:machine_type}

BARISTA spans three preparation styles---capsule, portafilter, and fully automatic---that differ in workflow complexity, object set, and interaction patterns.
We report per-task mean scores stratified by preparation type to verify whether these stylistic differences surface in model performance.

\begin{figure}[h]
    \centering
    \includegraphics[width=\linewidth]{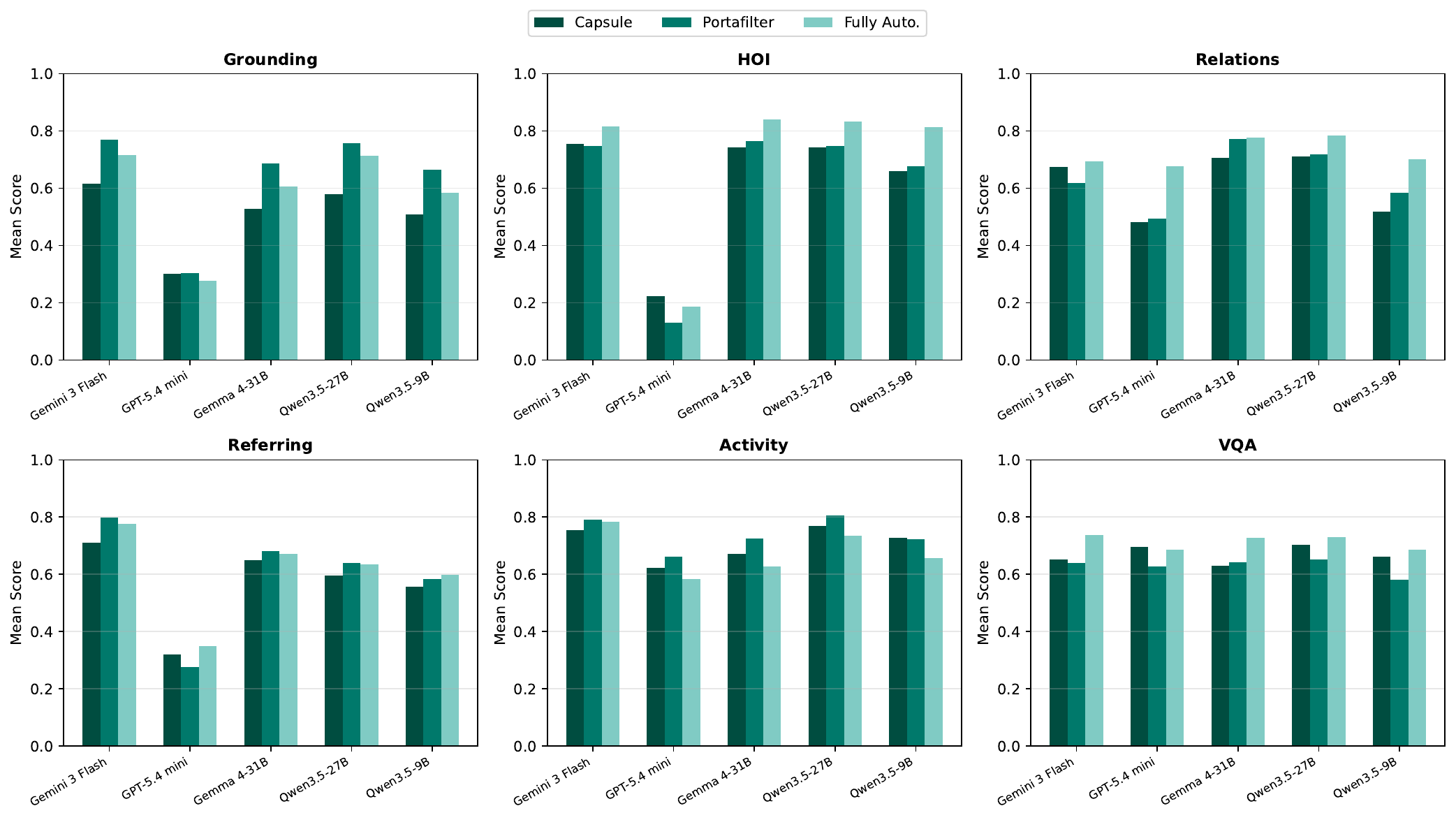}
    \caption{\textbf{Per-machine-type breakdown.} Mean per-example score by preparation style for each task and model. The three preparation styles expose different performance profiles: portafilter excels in grounding and activity recognition, while fully automatic dominates relation extraction and HOI.}
    \label{fig:machine_type}
\end{figure}

\paragraph{Results.}
Table \ref{tab:machine_type} and Figure \ref{fig:machine_type} report the full breakdown.
Performance varies meaningfully across preparation types in a task-dependent manner.
Portafilter videos yield the highest grounding scores (mean $0.636$ vs.\ $0.507$ for capsule), likely because the portafilter workflow involves larger, more distinctive objects at closer range.
Conversely, fully automatic machines produce the best scores on relation extraction ($0.727$ vs.\ $0.619$ capsule) and HOI ($0.698$ vs.\ $0.614$ portafilter), reflecting a simpler, more constrained interaction graph.
Activity recognition is easiest for portafilter ($0.742$) yet hardest for fully automatic ($0.677$), suggesting that models can recognize the distinctive manual steps (tamping, locking) more reliably than the subtler stages of an automated cycle.
These divergent patterns confirm that the three preparation styles expose complementary performance characteristics rather than uniformly scaling difficulty.

\begin{table}[h]
\centering
\small
\caption{\textbf{Per-machine-type mean scores.} Average per-example score across all five models, stratified by preparation style and task (scale $[0,1]$, higher is better). Scores are: mean best IoU (Grounding), interaction $F_1$ (HOI), tuple $F_1$ (Relations), G-Eval mean (Referring, VQA), accuracy (Activity).}
\label{tab:machine_type}
\setlength{\tabcolsep}{5pt}
\begin{tabular}{@{}l cccccc@{}}
\toprule
Preparation & Grounding & HOI & Relations & Referring & Activity & VQA \\
\midrule
Capsule        & $0.507$ & $0.624$ & $0.619$ & $0.567$ & $0.709$ & $0.669$ \\
Portafilter    & $0.636$ & $0.614$ & $0.638$ & $0.595$ & $0.742$ & $0.628$ \\
Fully Auto.    & $0.579$ & $0.698$ & $0.727$ & $0.606$ & $0.677$ & $0.714$ \\
\bottomrule
\end{tabular}
\end{table}

\newpage
\section{Input-ablation controls for activity recognition}
\label{app:input_controls}

To test how much activity recognition relies on visual evidence and temporal order, we rerun Qwen3.5-27B on the same $500$ clips with four input controls: first and last frame only, all four frames in random order, the central frame only, and text only (question and $25$ choices).
We compare these with the full input of four ordered frames, keeping the prompts and decoding unchanged.
Each condition is run three times, with no invalid responses.
We report accuracy differences with $95\%$ confidence intervals from a video-clustered paired bootstrap ($2{,}000$ resamples across $134$ source videos).

All four controls significantly reduce accuracy (Table~\ref{tab:input_controls}).
Text-only accuracy is $8.6\%$, compared with $4\%$ chance and $76.6\%$ with full input, indicating limited performance from language priors alone.
Removing intermediate frames lowers accuracy by $4.3$\,pp, shuffling by $9.7$\,pp, and using a single frame by $21.0$\,pp.
These results show that both visual evidence and temporal order contribute to activity recognition.

\begin{table}[h]
\centering
\small
\caption{\textbf{Input-ablation controls} for activity recognition with Qwen3.5-27B on the frozen $500$ clips (mean $\pm$ std over three runs; $\Delta$ in pp with video-clustered paired bootstrap $95\%$ CIs).}
\label{tab:input_controls}
\setlength{\tabcolsep}{6pt}
\begin{tabular}{@{}l c c@{}}
\toprule
Condition & Accuracy & $\Delta$ vs.\ full (pp, $95\%$ CI) \\
\midrule
Full ($4$ frames) & $.766 \pm .007$ & -- \\
Endpoints only    & $.723 \pm .008$ & $-4.3\ [-6.9, -1.8]$ \\
Shuffled          & $.669 \pm .013$ & $-9.7\ [-13.2, -6.4]$ \\
Single frame      & $.556 \pm .005$ & $-21.0\ [-25.1, -16.8]$ \\
Text only         & $.086 \pm .017$ & $-68.0\ [-72.2, -63.6]$ \\
\bottomrule
\end{tabular}
\end{table}



\end{document}